\documentclass[10pt,twocolumn,letterpaper]{article}

\usepackage{cvpr}              

\usepackage[utf8]{inputenc} 
\usepackage[T1]{fontenc}    
\usepackage{url}            
\usepackage{booktabs}       
\usepackage{amsmath,amssymb,amsfonts,amsthm}
\usepackage{nicefrac}       
\usepackage{microtype}      
\usepackage{xcolor}         

\usepackage{mathtools}
\usepackage{array}
\usepackage{ragged2e}

\makeatletter
\@namedef{ver@everyshi.sty}{}

\usepackage{tikz}
\usetikzlibrary{patterns}
\usetikzlibrary{spy}
\usepackage{color, colortbl}
\usepackage{pgfplots}
\pgfplotsset{compat=newest}
\usetikzlibrary{shapes.geometric}

\usepackage{multicol}
\usepackage{circuitikz}
\usepackage[nolist]{acronym}
\usepgfplotslibrary{groupplots}
\usepackage{graphicx}

\definecolor{mittelblau}{RGB}{0, 126, 198}
\definecolor{violettblau}{cmyk}{0.9, 0.6, 0, 0}
\definecolor{rot}{RGB}{238, 28 35}
\definecolor{apfelgruen}{RGB}{140, 198, 62}
\definecolor{gelb}{RGB}{1, 221, 0}
\definecolor{orange}{RGB}{244, 111, 33}
\definecolor{pink}{RGB}{237, 0, 140}
\definecolor{lila}{RGB}{128, 10, 145}
\definecolor{hellgrau}{RGB}{224, 224, 224}
\definecolor{mittelgrau}{RGB}{128, 128, 128}
\definecolor{dunkelgrau}{RGB}{80,80,80}
\definecolor{anthrazit}{RGB}{19, 31, 31}
\definecolor{darkgreen}{RGB}{0.125,0.5,0.169}

\usepackage{multirow}
\usepackage{float}

\newcolumntype{Q}{>{\raggedright\arraybackslash}p{0.6\textwidth}}
\newcolumntype{A}{>{\columncolor{gray!20!white}}p{0.4\textwidth}}
\newcolumntype{C}{c}
\newcolumntype{B}{>{\columncolor{gray!15!white}}c}
\newcolumntype{H}{>{\columncolor{gray!15!white}}r}
\newcolumntype{I}{>{\columncolor{gray!15!white}}l}
\newcolumntype{E}{>{\raggedright\arraybackslash}p{0.25\textwidth}}
\newcolumntype{J}{>{\raggedright\arraybackslash}p{0.125\textwidth}}
\newcolumntype{L}{>{\raggedright\arraybackslash}p{0.28\textwidth}}
\newcolumntype{K}{>{\raggedright\arraybackslash}p{0.01\textwidth}}
\newcolumntype{R}{>{\raggedleft\arraybackslash}p{0.33\textwidth}}
\usepackage{longtable}

\usepackage{caption}
\usepackage{enumitem}
\usepackage{wrapfig}
\usepackage{pifont}
\newcommand{\cmark}{\ding{51}}%
\newcommand{\xmark}{\ding{55}}%

\newcommand{\dataset}{MaxRay}
\newcommand{\raycast}{MaxRay}

\newcommand*\circled[1]{\tikz[baseline=(char.base)]{
            \node[shape=circle,fill,inner sep=1.2pt] (char) {\textcolor{white}{#1}};}}
            
\newcommand*{\inlineequation}[2][]{%
  \begingroup
    \refstepcounter{equation}%
    \ifx\\#1\\%
    \else
      \label{#1}%
    \fi
    \relpenalty=10000 %
    \binoppenalty=10000 %
    \ensuremath{%
      #2%
    }%
    ~\@eqnnum
  \endgroup
}

\DeclareMathSizes{10}{8}{7}{6}

\usepackage[pagebackref,breaklinks,colorlinks]{hyperref}

\usepackage[capitalize]{cleveref}
\crefname{section}{Sec.}{Secs.}
\Crefname{section}{Section}{Sections}
\Crefname{table}{Table}{Tables}
\crefname{table}{Tab.}{Tabs.}

\def\cvprPaperID{6019} 
\def\confName{CVPR}
\def\confYear{2023}

\usepackage[toc,page]{appendix}

\begin{document}
\begin{multicols}{2}
\begin{acronym}[WSSUS]

    \acro{3GPP}{3rd Generation Partnership Project}
    \acro{5G}{fifth-generation}
    \acro{ADC}{analog to digital converter}
    \acro{AFE}{analog front end}
    \acro{AGC}{automatic gain control}
    \acro{AGV}{automated guided vehicle}
    \acro{AMP}{approximate message passing}
    \acro{API}{Application Programming Interface}
    \acro{AWGN}{additive white Gaussian noise}
    \acro{AutoML}{automatic-machine-learning}
    \acro{BER}{bit error rate}
    \acro{BB}{baseband}
    \acro{bpcu}{bits per channel use}
    \acro{BP}{belief propagation}
    \acro{BPSK}{binary phase shift keying}
    \acro{BS}{base station}

    \acro{CFAR}{Constant-False-Alarm-Rate}
	\acro{CL}{Constrastive Loss}
	\acro{MCL}{Masked contrastive Loss}
    \acro{CB}{codebook}
    \acro{CDF}{cumulative distribution function}
    \acro{CFO}{carrier frequency offset}
    \acro{CoSaMP}{compressive sampling matching pursuit}
    \acro{CP}{cyclic prefix}
    \acro{CS}{compressive sensing} 
    \acro{CSI}{channel state information}
    \acro{CNN}{convolutional neural network}

    \acro{DA}{domain adaptation}
    \acro{DBSCAN}{Density-Based Spatial Clustering of Applications with Noise}
    \acro{DAC}{digital-analog-converter}
    \acro{DC}{direct current}
    \acro{DE}{distance error}
    \acro{DeepL}{deep-learning}
    \acro{DoF}{degree-of-freedom}
    \acro{DFT}{discrete Fourier transformation}
    \acro{DL}{deep learning}
    \acro{DS}{delay spread}
    \acro{DSP}{digital signal processing}

    \acro{ECC}{error-correcting code}
    \acro{ENoB}{effective number of bits}
    \acro{ERP}{effective radiated power}
    \acro{EVM}{error vector magnitude}
    \acro{EVD}{eigenvector decomposition}
    \acro{FB}{feedback}
    \acro{FC}{fully connected}
    \acro{FDD}{frequency division duplexing}
    \acro{FDM}{frequency division multiplexing}
    \acro{FIR}{finite impulse response}
    \acro{FFT}{fast fourier transform}
    \acro{FT}{fine tuning}
    \acro{FPGA}{field programmable gate array}
    \acro{GAN}{Generative adversarial network}
    \acro{GPIO}{general-purpose input/output}
    \acro{GPS}{global positioning system}
    \acro{GPSDO}{GPS disciplined oscillator}
    \acro{GPU}{graphical processing unit}
    \acro{HDF}{Hierarchical Data Format}
    \acro{HDD}{hard decision decoding}
    \acro{IC}{integrated circuit}
    \acro{ICI}{inter-carrier-interference}
    \acro{ISAC}{Integrated Sensing And Communication}
    \acro{I2C}{Inter-Integrated Circuit}
    \acro{ICSP}{in-circuit serial programming}
    \acro{IF}{intermediate frequency}
    \acro{i.i.d.}{independent and identically distributed}
    \acro{IIR}{infinite impulse response}
    \acro{IMU}{inertial measurement unit}
    \acro{IoT}{Internet of Things}
    \acro{IPS}{indoor positioning system}
    \acro{IR}{infrared}
    \acro{JSDM}{Joint Spatial Division and Multiplexing}
    \acro{LIDAR}{Light Detection And Ranging}
    \acro{LLR}{log-likelihood ratio}
    \acro{LP}{leakage precoder}
    \acro{LMMSE}{Linear Minimum Mean Square Error}
    \acro{LO}{local oscillator}
    \acro{LoS}{line of sight}
    \acro{LiDaR}{Light Detection and Ranging}
    \acro{LS}{least squares}
    \acro{LSTM}{long-term short-term memory}
    \acro{LTE}{Long Term Evolution}
    \acro{LTI}{linear time invariant}
    \acro{LTV}{linear time variant}
  
    \acro{MAP}{maximum a posteriori}
    \acro{MDE}{mean distance error}
    \acro{MDA}{mean distance accuracy}
    \acro{MEMS}{Micro-Electro-Mechanical Systems}
    \acro{MIMO}{multiple input multiple output}
    \acro{MISO}{multiple input single output}
    \acro{ML}{maximum likelihood}
    \acro{MLD}{maximum likelihood decoding}
    \acro{mMIMO}{massive multiple input multiple output}
    \acro{MMSE}{minimum mean square error}
    \acro{M-MMSE}{multi-cell minimum mean square error}
    \acro{MR}{maximum ratio}
    \acro{MRC}{maximum ratio combining}
    \acro{MRP}{maximum ratio precoding}
    \acro{MRT}{maximum ratio transmission}
    \acro{MSE}{mean squared error}
    \acro{MQTT}{Message Queuing Telemetry Transport}
    \acro{MU}{multi-user}
    \acro{MUSIC}{Multiple Signal Classification}
    \acro{NF}{noise figure}
    \acro{NN}{Neural Network}
    \acro{NNI}{Neural Network Intelligence}
    \acro{NLoS}{non-line of sight}
    \acro{NND}{neural network decoding}
    \acro{NTP}{Network Time Protocol}
    \acro{NMSE}{normalized mean squared error}
    \acro{NU}{not-used}
    \acro{OFDM}{orthogonal frequency division multiplex}
    \acro{OMP}{orthogonal matching pursuit}
    \acro{OPS}{outdoor positioning system}
    \acro{OT}{optimal transport}
    \acro{PB}{passband}
    \acro{PCB}{printed circuit board}
    \acro{PDR}{pedestrian dead reckoning}
    \acro{PDF}{probability density function}
    \acro{PDP}{power-delay-profile}
    \acro{PLL}{phase-locked-loop}
    \acro{PO}{phase-only}
    \acro{PPS}{pulse per second}
    \acro{QPSK}{quadrature phase shift keying}
    \acro{QuaDRIGa}{Quasi Deterministic Radio Channel Generator}

    \acro{RADAR}{Radio Detection And Ranging}
    \acro{ReLU}{rectified linear unit}
    \acro{RF}{radio frequency}
    \acro{RMS-DS}{Root Mean Square - Delay Spread}
    \acro{RNN}{recurrent neuronal network}
    \acro{RSSI}{received signal strength indicator}
    \acro{R-ZF}{regularized zero-forcing}
    \acro{SDD}{soft decision decoding}
    \acro{SDR}{software defined radio}
    \acro{SE}{spectral efficiency}
    \acro{SFO}{sampling frequency offset}
    \acro{STO}{sampling time offset}
    \acro{SLAM}{Simultaneous Localization and Mapping}
    \acro{SGD}{stochastic gradient descent}
    \acro{SISO}{single input single output}
    \acro{SINR}{signal-to-interference-and-noise-ratio}
    \acro{SIR}{signal-to-interference-ratio}
    \acro{SLNR}{signal-to-leakage-and-noise ratio}
    \acro{SNR}{signal-to-noise-ratio}
    \acro{SP}{subspace}
    \acro{SQR}{signal-to-quantization-noise-ratio}
    \acro{SQNR}{signal-to-quantization-noise-ratio}
    \acro{SVD}{singular value decomposition}
    \acro{SU}{single-user}
    \acro{TDD}{time division duplexing}
    \acro{TRIPS}{time-reversal IPS}
    \acro{UE}{user equipment}
    \acro{UL}{uplink}
    \acro{ULA}{uniform line array}
    \acro{URLLC}{ultra-reliable low-latency communication}
    \acro{US}{uncorrelated scattering}
    \acro{USRP}{universal software radio peripheral}
    \acro{UWB}{ultra-wideband}
    \acro{WiFi}{Wireless Fidelity}
    \acro{WSS}{wide sense stationary}
    \acro{WSSUS}{wide sense stationary uncorrelated scattering}

    \acro{ZF}{zero forcing}
\end{acronym}
\end{multicols}

\title{Look, Radiate, and Learn:\\ Self-Supervised Localisation via Radio-Visual Correspondence}

\author{%
  Mohammed~Alloulah\thanks{Correspondence to \texttt{alloulah@outlook.com}}
  \quad\quad
  Maximilian~Arnold\thanks{Work done whilst at Nokia Bell Labs.}
  \\
  \vspace{0.6em}
  Nokia Bell Labs
}

\maketitle

\begin{abstract}
\vspace{-0.20cm}
  Next generation cellular networks will implement radio sensing functions alongside customary communications, thereby enabling unprecedented worldwide sensing coverage outdoors. Deep learning has revolutionised computer vision but has had limited application to radio perception tasks, in part due to lack of systematic datasets and benchmarks dedicated to the study of the performance and promise of radio sensing.
  To address this gap, we present~\dataset: a synthetic radio-visual dataset and benchmark that facilitate precise target localisation in radio. We further propose to learn to localise targets in radio without supervision by extracting self-coordinates from radio-visual correspondence. We use such self-supervised coordinates to train a radio localiser network. We characterise our performance against a number of state-of-the-art baselines. Our results indicate that accurate radio target localisation can be automatically learned from paired radio-visual data without labels, which is important for empirical data. This opens the door for vast data scalability and may prove key to realising the promise of robust radio sensing atop a unified communication-perception cellular infrastructure. Dataset will be hosted on IEEE DataPort.
\end{abstract}

\vspace{-0.50cm}
\section{Introduction} \label{sec:intro}

\begin{figure}
  \centering
  \includegraphics[width=1.00\columnwidth]{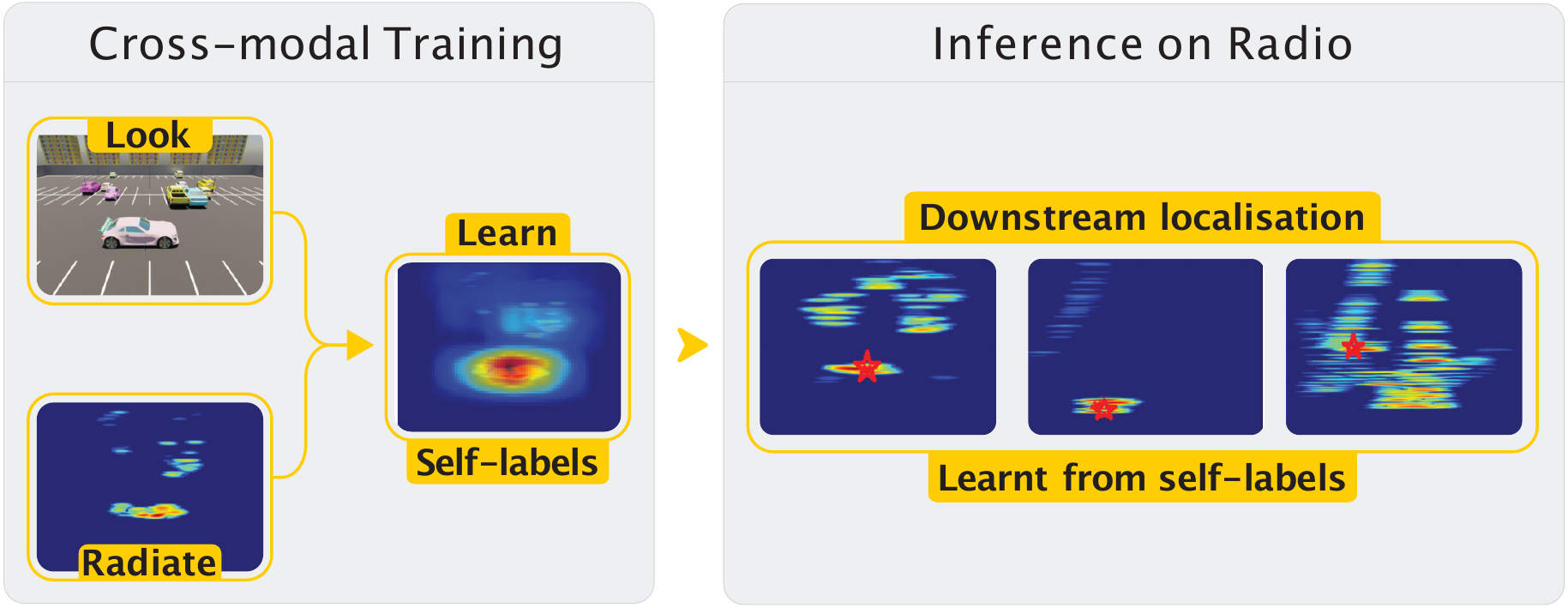}
  \vspace{-0.75cm}
  \caption{\small We train a radio localisation network by using commonalities with vision to drive spatial attention. Without laborious manual annotations, we learn to suppress clutter and localise targets in radio heatmaps.}
\label{fig:abstractoverview}
\end{figure}

Sixth-generation (6G) wireless networks are being designed from the ground up to support sensing at the physical layer~\cite{wild2021joint}.
Such a brand new capability in 6G networks marks a departure from communication-only functions, and aims to supply applications with sensing primitives atop a unified communication-perception infrastructure.
Concretely, dense cellular deployments in urban settings (e.g., per lamppost) would allow for unprecedented radio coverage, enabling a multitude of challenging perception tasks.
Examples include around-the-corner obstacle detection in support of autonomous driving and pedestrian and drone localisation, to name a few~\cite{bourdoux20206g}.

Training perception models for radio signals is a key challenge for network infrastructure vendors.
Unlike vision and audio, radio signals are hard to label manually because they are not human interpretable.
Typically, sparse radio signals have been paired with a groundtruth vision modality for reliable semantic and qualitative filtration via a cross-modal annotation flow~\cite{wang2021rodnet,zhao2018through,li2019making,fan2020learning}.
Recently, this radio-visual pairing has been shown to work in a self-supervised fashion~\cite{alloulah2021self}, building on a wave of progress in vision self-supervised learning (SSL)~\cite{chen2020simple,he2020momentum,grill2020bootstrap,zbontar2021barlow,noroozi2016unsupervised,zhang2016colorful,gidaris2018unsupervised,oord2018representation,caron2020unsupervised,bardes2021vicreg,goyal2022vision}.  

Computer vision has traditionally benefited from synthetic datasets for: (a) content augmentation for enhanced generalisability~\cite{krahenbuhl2018free,yao2020simulating}, or (b) closing the learning loop on out-of-distribution failure modes~\cite{risi2020increasing}, e.g., in the context of autonomous driving~\cite{TeslaAIDay21_Simulation}.
Extrapolating from vision, it is also likely that synthetic data will play an important role towards realising robust radio sensing.
However, radio perception tasks have yet to benefit from such publicly available datasets.

In this work we aim to support next-gen 6G perception tasks, while championing a self-supervised radio-visual learning approach.
Concretely, Fig.~\ref{fig:abstractoverview} captures the crux of our new machine learning proposition for radio sensing. 
We demonstrate how to automatically extract radio self-labels through cross-modal learning with vision. 
We then use such self-labels to train a downstream localiser network. 
We show that our self-supervised localiser net enhances estimation in the radio domain compared to state-of-the-art.
Our contributions are:
\begin{itemize}[noitemsep,nosep]
  \item A synthetic dataset: We curate and synthesise radio-visual data for a new learning task designed for target detection and localisation in radio.
  \item A cross-modal SSL algorithm: We formulate a contrastive radio-visual objective for label-free radio localisation.
  \item Evaluation: We conduct numerous characterisations on synthetic and empirical data in order to validate our SSL algorithm and expose its superior performance compared to state-of-the-art.
 \end{itemize}
We discuss our dataset and algorithmic findings to galvanise machine learners' interest in radio-visual learning research.
We hope to both facilitate and inform future research on this new cross-modal learning paradigm.

\vspace{-0.20cm}
\section{Related Work} \label{sec:related_work}

\noindent \textbf{Self-supervised learning.} Self-supervised learning (SSL) in its two strands (contrastive and non-contrastive) is the state-of-the-art learning paradigm for visual representations~\cite{chen2020simple,he2020momentum,grill2020bootstrap,zbontar2021barlow}. 
SSL models have progressively matched and then exceeded the performance of their fully-supervised counterparts~\cite{noroozi2016unsupervised,zhang2016colorful,gidaris2018unsupervised,oord2018representation,hjelm2018learning,bachman2019learning,caron2020unsupervised,henaff2020data,bardes2021vicreg}, culminating recently in strong performance on uncurated billion-scale data~\cite{goyal2022vision}.
Vision SSL relies on augmentation for semantic invariance.
Differently, we deal with a new radio-visual SSL problem that relies on cross-modal correspondence~\cite{arandjelovic2017look,arandjelovic2018objects} as opposed to augmentation.
Further, our work addresses SSL object detection and localisation using spatial backbone models~\cite{chatfield2014return,afouras2021self} rather than the prevalent object classification in vision using 1-D backbones.

\noindent \textbf{Self-supervised multi-modal object detection.} A related body of work leverages multiple modalities for representation learning, particularly between audio and vision~\cite{alwassel2020self,arandjelovic2017look,asano2019self,aytar2016soundnet,morgado2020learning,owens2016ambient,afouras2020self,arandjelovic2018objects,chen2021localizing}. 
Other works, also audio-visual, deal with knowledge distillation from one modality to another~\cite{gan2019self,afouras2020asr}.
SSL audio-visual object detectors are well researched and rely on feature attention between 1-D audio and 2-D vision~\cite{afouras2020self,afouras2021self}. 
Differently in radio-visual, our attention (a) is complicated by a sparse radio modality which could impact the dimensional stability of cross-modal contrastive learning~\cite{jing2021understanding}, and (b) involves a fundamentally larger feature search space between 2-D radio and 2-D vision.

\noindent \textbf{Self-supervised saliency localisation.} Recent works have extended visual saliency localisation~\cite{zhou2016learning,selvaraju2017grad} for self-supervised systems~\cite{baek2020psynet}.
Specifically, \cite{mo2021object}~expands class activation map (CAM) to work within an SSL network to markedly improve visual contrastive learning and mitigate against augmentation bias.
While notable for vision SSL, radio-visual SSL does not suffer from the augmentation-induced geometric perturbations during training (e.g., random crop and rotation) which make accurate object localisation trickier in vision SSL.

\noindent \textbf{Self-supervision with priors.} Some works bake prior information back into SSL, e.g., using off-the-shelf image segmentation models~\cite{henaff2020data,henaff2021efficient,van2021unsupervised}.
Follow-up works replace these priors with online learning that works hand in hand with SSL~\cite{henaff2022object,chen2021multisiam}.
Our work uses priors from vision to bootstrap radio-visual SSL in a relatively small data regime.
Similarly, however, radio-visual SSL could be made to work without vision priors in principle.

\noindent \textbf{Radio learning.} 
Recent works train radio models on vision-supplied labels for indoor and outdoor sensing, e.g.,~\cite{zhao2018through,li2019making,fan2020learning,guan2020through}.
SSL has also been recently applied to radio-only learning systems. 
\cite{orr2021coherent}~proposes an SSL super-resolution method that improves the angular resolution of radar antenna arrays. 
\cite{gasperini2021r4dyn}~uses radar during training as a weak supervision signal, as well as an extra input to enhance depth estimation at inference time.
\cite{li2022unsupervised}~tackles the problem of radio-only SSL for human sensing.
Our work is different from the above prior art in that it neither relies on explicit supervision from vision, nor it is single-modal for radio-only learning.
A recent work proposes radio-visual SSL for object classification within a distillation framework~\cite{alloulah2021self}.
This differs from our work which (a) deals with representation learning from scratch for both radio and vision and (b) is aimed at SSL object detection and localisation using an underlying spatial backbone as opposed to standard classification.

\noindent \textbf{Radio-visual datasets.} 
A number of multi-modal datasets (with radio-visual entries) are available in the adjacent automotive literature, where it is not uncommon for datasets that are collected using fleets of cars to be quite large.
Examples include CRUW~\cite{wang2021rodnet}, Carrada~\cite{ouaknine2021carrada}, AIODrive~\cite{Weng2020_AIODrive}, RADIATE~\cite{sheeny2021radiate}, 
Oxford Radar RobotCar~\cite{barnes2020oxford}.
\begingroup
\setlength{\columnsep}{6pt}%
\begin{wrapfigure}[12]{r}{.265\textwidth}
    \vspace{-0.35cm}
    \centering
    \captionsetup{type=table}
    \caption{\small Radio-visual datasets.$^{\dagger}$}
    \label{tab:rv_datasets_summary}
    \medskip
  \vspace{-0.50cm}
  \scriptsize
  \setlength{\tabcolsep}{2pt}
  \begin{tabular}{lcc}
    \toprule
    \multicolumn{1}{c}{Dataset}        & \multicolumn{1}{c}{Automotive}  & \multicolumn{1}{c}{6G}  \\
    \midrule
    CRUW~\cite{wang2021rodnet}         & \cmark                          &  \xmark                 \\
    Carrada~\cite{ouaknine2021carrada} & \cmark                          &  \xmark                 \\
    AIODrive~\cite{Weng2020_AIODrive}  & \cmark                          &  \xmark                 \\
    RADIATE~\cite{sheeny2021radiate}   & \cmark                          &  \xmark                 \\
    Oxford Radar RobotCar~\cite{barnes2020oxford} & \cmark                          &  \xmark                 \\
    RADDet~\cite{zhang2021raddet}      & \xmark                          &  \cmark                 \\
    DeepSense~\cite{DeepSense}         & \xmark                          &  \cmark                 \\[-0.05cm]
    \midrule \\[-0.4cm]
    \dataset$^{\ast}$                  & \xmark                          &  \cmark                 \\
    \bottomrule             
  \end{tabular}
  \scriptsize{
  \parbox{.26\textwidth}{$^{\ast}$\dataset~is the only 6G synthetic dataset.} \\
  \parbox{.26\textwidth}{$^{\dagger}$Refer to Tab.~\ref{tab:rv_datasets_verbose} in Appendix~\ref{sec:cruw} for a more detailed comparison.}
  }
\end{wrapfigure} 
6G networks, on the other hand, focus on radio-visual data collected at a stationary basestation for sensing the surrounding environment. 
RADDet~\cite{zhang2021raddet} and DeepSense~\cite{DeepSense} are closely related datasets. 
However, both are empirical datasets with low angular resolution. 
In contrast, our synthetic dataset has higher angular resolution and incorporates high-fidelity propagation modelling\footnote{made possible by decades of statistical radio modelling~\cite{zwick2002stochastic,fishler2006spatial}} and graphical rendering, which result in quality radio-visual data.
This allows for much tighter characterisation and refinement of algorithms given the (a) controllability (e.g., configurability w.r.t. radio parameters, cf. Tab.~\ref{tab:radio_params}) and (b) measurability against perfect groundtruth.

\endgroup

\vspace{-0.25cm}
\section{Dataset} \label{sec:dataset}

\begin{figure}[h]
  \vspace{-0.45cm}
    \centering
    \caption{\small Block diagram of~\raycast.}
    \medskip
    \vspace{-0.50cm}
    \small
    \includegraphics{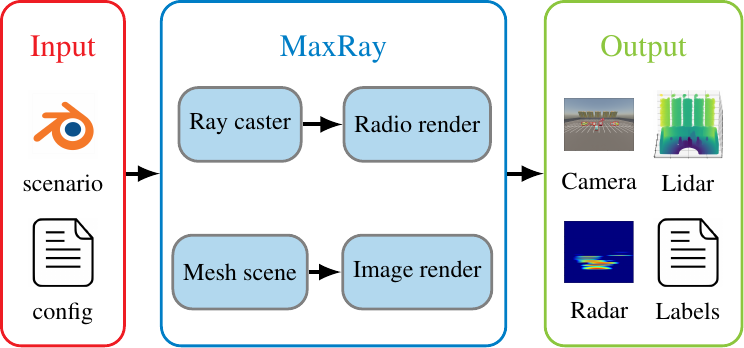}
    \label{fig:framework}    
\end{figure}

\vspace{-0.6cm}
\begin{table}[h]
\begin{minipage}[t]{.260\textwidth}
    \centering
    \caption{\small Sensor entries in~\dataset.}
    \label{tab:dataset}
    \medskip
    \vspace{-0.5cm}
  \scriptsize
  \setlength{\tabcolsep}{2pt}
  \begin{tabular}{lll}
    \toprule
    \multicolumn{1}{c}{Entry}   & \multicolumn{1}{c}{Type}     & \multicolumn{1}{c}{Label}     \\
    \cmidrule(lr){1-3}
    Camera      & Image                     & bounding box + class                                  \\
    Lidar       & Tensor                    & per point material + class                            \\
    Depth       & Image                     & bounding box + class                                  \\
    Radar       & Image                     & bounding box + reflectors + class                     \\
    CSI         & Tensor                    & reflectors + paths AoA$^{\ast}$/AoD$^{\dagger}$       \\
    \bottomrule             
  \end{tabular}
  \scriptsize{
  \parbox{1\textwidth}{$^{\ast}$AoA: angle of arrival} \\
  \parbox{1\textwidth}{$^{\dagger}$AoD: angle of departure}
  }
\end{minipage}\hfill
\begin{minipage}[t]{.185\textwidth}
  \centering
  \caption{\small Radio synthesis.}
  \label{tab:radio_params}
  \medskip
  \vspace{-0.5cm}
  \scriptsize
  \setlength{\tabcolsep}{1pt}
  \begin{tabular}{lc}
    \toprule
    Parameter           & Value           \\
    \cmidrule(lr){1-2} 
    RX array            & 16$\times$16    \\
    Bandwidth           & 800MHz          \\
    Carrier             & 28GHz           \\
    \multirow{2}{*}{\parbox{1.5cm}{Range-Angle Bins}}  & \multirow{2}{*}{480$\times$640}  \\
                        &                 \\
    \bottomrule             
  \end{tabular}
\end{minipage}
\end{table}

\vspace{-0.25cm}
Our radio-visual dataset is created using~\raycast~\cite{arnold2022maxray}---a ray tracing tool for accurate radio propagation simulations.
MaxRay also incorporates the open-source Blender engine for creating photo-realistic environments~\cite{Blender_sw}. 
As such, we can model arbitrarily complex environments and synthesise paired responses in the vision and radio domains.

Fig.~\ref{fig:framework} depicts the tool block diagram.
A Blender scenario and a configuration file (containing radio parameters such as carrier frequency and bandwidth) are inputted to~\raycast. 
\raycast~uses Python APIs to render responses for a variety of imaging sensors (e.g., camera, lidar, depth images) along with their labels. 
The rendering and label quality allow us to train an off-the-shelf Yolo v5 models~\cite{glenn_jocher_2021_5563715} from scratch.
The core of~\raycast~leverages the ray casting capability of Blender to simulate complex radio phenomena (e.g., scattering and reflection) and calculate their propagation losses. 
These propagation losses are then used to create channel state information (CSI), which is in turn converted to radar heatmaps according to an orthogonal frequency-division multiplexing (OFDM) signalling architecture.

\subsection{Modelling \& synthesis details} \label{sec:dataset_modelling_details}

\vspace{-0.1cm}
\noindent \textbf{Vision.} 
We model everything in Blender. 
Currently, we implement five different materials (glass, wood, concrete, metal, and water), 20 different building types, and 28 unique and accurate car models. 
Once the dataset is open-sourced, the research community can build on our Blender models to further extend the scale and richness of our radio-visual dataset.\footnote{See datasheet in Appendix~\ref{sec:datasheet} for further details on extension.} 
As of now, we sample from a standard normal distribution (10cm standard deviation) to randomise the location of 28 unique cars on the road, while also permuting their colour.
The pose is constrained by the lane along which cars are travelling. 
Different scenarios are evenly distributed.

\noindent \textbf{Radio.}
We generate radio heatmaps that correspond to visual Blender models using ray tracing.
Appendix~\ref{sec:appendix_ofdm_radar} treats the signal processing principles of OFDM radar, which we implement in our synthesis flow.
We validate our ray tracing against empirical measurements, as well as other commercial ray tracers.
Recent analysis against~\cite{DeepSense} reveals effective behavioural modelling.
Our current scenarios focus on cars.
We plan to extend to scenarios featuring humans, which require elaborate modelling of micro Doppler effects~\cite{alloulah2019kinphy}. 

\vspace{-0.20cm}
\subsection{Version 1.0}

\begin{figure}[t]
    \centering
    \includegraphics{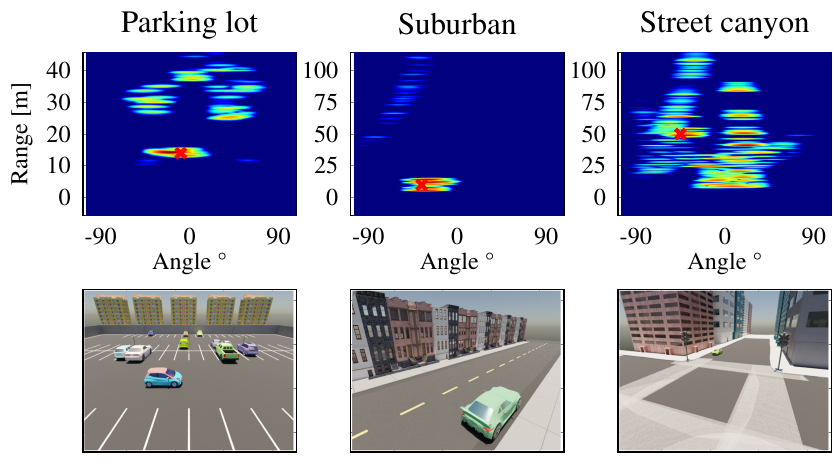}
    \vspace{-0.85cm}
    \caption{\small Example radio-visual heatmap-image pairs of three different scenarios: parking lot (left), suburban (middle), and street canyon (right)}
    \label{fig:example_radio-visual_pairs} 
\end{figure}

\begin{figure*}
  \centering
  \includegraphics[width=1.0\textwidth]{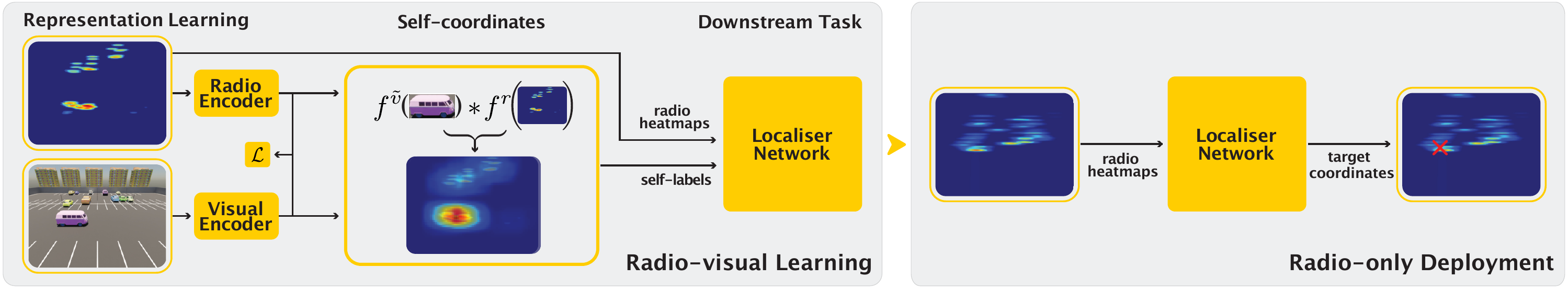}
  \vspace{-0.75cm}
  \caption{\small Radio target localisation via self-supervised radio-visual correspondence. We combine contrastive radio-visual learning with visual masking to extract radio target self-coordinates, on which we then train a radio-only localiser net.}
  \label{fig:radio-visual_ssl}
\end{figure*}

\vspace{-0.1cm}
The current version of the dataset supports the sensor configurations listed in Tab.~\ref{tab:dataset}. 
All available sensors are paired and synchronised per data point, facilitating cross-modal learning.
Tab.~\ref{tab:radio_params} lists the radio configurations used in dataset, which comply with current 5G Advanced specifications~\cite{kim2019new}.
Further, the dataset has sequences of 15 data points that allow for time series modelling.

The full version of the dataset has 3 scenarios: a parking lot, a suburban street, and a street canyon.
Fig.~\ref{fig:example_radio-visual_pairs} depicts one example per scenario. 
In parking lot, one car is driven from left to right or right to left. 
In suburban, one car drives along the houses towards the camera or away from it. 
The same holds for the street canyon scenario.
Note how radio heatmaps have different ranges, as well as different amount of spurious clutter.  
For instance, parking lot has dynamic background clutter arising from changes in the location and pose of stationary cars across data points.
Groundtruth information is supplied in the form of bounding boxes for vision and target polar coordinates for radar (i.e., range and angle).
In the terms of data diversity, there are 50 different cars, backgrounds and foregrounds randomised throughout the dataset. 
There are also portions of data that model mixed weather scenarios such as rain, snow, fog, and dust.
We provide a few dataset illustrations in Appendix~\ref{sec:appendix_maxray_illustra}.

Phase 1 of dataset release focuses on the parking lot scenario only.
For the remainder of this paper, we use parking lot with radar and camera entries.
Specifically, parking lot has 30,000 paired radio-visual data points, split into 24k training and 6k validation sets.
An additional 10k set is withheld for testing.
All our results are reported on the 6k validation set.

\vspace{-0.20cm}
\section{Method}

We aim to automatically localise a target of interest in radio by tapping into the common information radio and its paired vision capture about the physical world.
In fact, barring propagation nuances, radio imaging can be thought of as a low-resolution form of vision---Appendix~\ref{sec:radio-visual_analytic_relationship} justifies this view using a 1st-order analytic analysis derived from first principles.
As such, jointly embedding radio and vision becomes not only a convenience, but is also naturally grounded in physics.
Therefore, we would hope that the joint embedding architecture would constitute a powerful representation for building a wide variety of radio-only or combined radio-visual perception tasks.
Concretely, our approach in this paper is to: \circled{1} learn cross-modal spatial features via radio-visual correspondence, \circled{2} extract self-estimates of target coordinates (i.e., pseudo labels) via cross-modal attention between the spatial features, and \circled{3} use the self-coordinates to train a radio-only target localiser network. Fig.~\ref{fig:radio-visual_ssl} illustrates this three-step procedure.
In what follows, we explain further \circled{1},~\circled{2}, and~\circled{3}.

\vspace{-0.20cm}
\subsection{Representation learning} \label{sec:representation_learning}

\vspace{-0.1cm}
We employ a flavour of contrastive learning we dub masked contrastive learning (MCL) in order to self-localise targets in radio.
MCL is inspired by earlier pioneering audio-visual learning works~\cite{arandjelovic2018objects,arandjelovic2017look}, as well as canonical visual contrastive learning~\cite{chen2020simple,chen2020improved}.
We begin by formalising MCL.

\noindent \textbf{Masked contrastive learning (MCL).} \label{sec:scl}
Let $(r, v)$ be a radio-visual data pair, where $r \in \mathbb{R}^{1 \times H \times W}$ is a radar heatmap and $v \in \mathbb{R}^{3 \times H \times W}$ is a corresponding RGB image.
Encode, respectively, radio and vision by two backbone nets $f_{\theta^r}$ and $f_{\theta^v}$, and their momentum-filtered versions $f_{\bar{\theta}^r}$ and $f_{\bar{\theta}^v}$, assuming some weight parametrisation $\{\theta^r, \theta^v\}$.
Each backbone net encodes per bin one $C$-dimensional feature vector within 2-dimensional spatial bins, i.e., $f_{\theta^r}(r), f_{\theta^v}(v) \in \mathbb{R}^{C \times h \times w}$. 
The spatial binning resolution $h \times w$ is generally coarser than the original image resolution $H \times W$.
Denote by $f^{r}_{n}(r), f^{v}_{n}(v) \in \mathbb{R}^{C}$ radio and vision spatial encodings at bin $n \in \Omega = \{1, \dots, h\} \times \{1, \dots, w\}$.
Construct a visual target mask $\gamma \coloneqq [ \gamma_{ij} ] \in [0, 1]^{H \times W}$ such that $f^{\tilde{v}}_{m}(\gamma \odot v) \in \mathbb{R}^{C}$ is defined for $m \in \tilde{\Omega} = \{1, \dots, \tilde{h}\} \times \{1, \dots, \tilde{w}\}$ to retain encodings for the target of interest only in the RGB image (e.g., as delineated by a bounding box), where $\odot$ is the element-wise product, $\tilde{\Omega} \subset \Omega$ is a subset of spatial locations, and $\tilde{v}$ denotes masking in vision.
In practice, the target mask can either be (1) estimated using off-the-shelf vision object detectors such as Yolo~\cite{redmon2016you,glenn_jocher_2021_5563715}, or (2) obtained directly as groundtruth during data synthesis.
Use 2-layer MLP projector heads $g_{\theta^r}$ and $g_{\theta^v}$ to collapse the spatial encodings of the backbone nets $f_{\theta^r}$ and $f_{\theta^v}$ onto vector representations as
\vspace{-0.20cm}
\begin{align}
  q^r &= g_{\theta^r}(f_{\theta^r}(r)), \hspace{0.80cm} k^{\tilde{v}} = g_{\theta^v}(f_{\bar{\theta}^v}(\gamma \odot v)) \notag \\ 
  q^{\tilde{v}} &= g_{\theta^v}(f_{\theta^v}(\gamma \odot v)), \hspace{0.25cm} k^r = g_{\theta^r}(f_{\bar{\theta}^r}(r)) \notag 
\end{align}
where vectors $q^r, q^{\tilde{v}}, k^r, k^{\tilde{v}} \in \mathbb{R}^{N}$, superscripts $r$ and ${\tilde{v}}$ denote respectively radio and masked vision, and following MoCo's query $q$ and key $k$ notation~\cite{he2020momentum}.
With each $r$, use $K+1$ samples of ${\tilde{v}}$ of which one sample ${\tilde{v}}^{+}$ is a true match to $r$ and $K$ samples $\{{\tilde{v}}_{i}^{-}\}_{i=0}^{K-1}$ are false matches---vice versa with each ${\tilde{v}}$, $K+1$ samples of $r$.
The one-sided cross-modal contrastive losses that test for masked vision-to-radio and radio-to-masked vision correspondences are
\vspace{-0.20cm}
\begin{align}
  \mathcal{L}_c^{\tilde{v} \rightarrow r}(q^r, k^{{\tilde{v}}+}, \mathbf{k}^{{\tilde{v}}-}) &= - \underset{r,v}{\mathbb{E}} \log \frac{ e^{\operatorname{sim}(q^r, k^{{\tilde{v}}+})} }{ e^{\operatorname{sim}(q^r, k^{{\tilde{v}}+})} +\sum\nolimits_{i} e^{\operatorname{sim}(q^r, k_{i}^{{\tilde{v}}-})} } \notag \\
  \mathcal{L}_c^{r \rightarrow {\tilde{v}}}(q^{\tilde{v}}, k^{r+}, \mathbf{k}^{r-}) &= - \underset{r,v}{\mathbb{E}} \log \frac{ e^{\operatorname{sim}(q^{\tilde{v}}, k^{r+})} }{ e^{\operatorname{sim}(q^{\tilde{v}}, k^{r+})} +\sum_{ i } e^{\operatorname{sim}(q^{\tilde{v}}, k_{i}^{r-})} } \notag
\end{align}
where $\operatorname{sim}(x, y) \coloneqq x^\top y/ \tau$ is a similarity function, $\tau$ is a temperature hyper-parameter, $k^{x+/-} = g_{\theta^x}(f_{\bar{\theta}^x}(x^{+/-}))$ are encodings that denote true and false corresponding signals $x \in [r, {\tilde{v}}]$ , and vector $\mathbf{k}^{x-} = \{k_{i}^{x-}\}_{i=0}^{K-1}$ holds $K$ false encodings. 
Then the bidirectional masked contrastive loss\footnote{see Fig.~\ref{fig:contrastive_architectures} in Appendix~\ref{sec:cl_background} for further illustration} that incentivises cross-modal spatial attention becomes
\vspace{-0.25cm}
\begin{align}
  \mathcal{L}_{\text{MCL}} &= ( \mathcal{L}_c^{\tilde{v} \rightarrow r} + \mathcal{L}_c^{r \rightarrow \tilde{v}} )/2
\end{align}

\vspace{-0.20cm}
After training, the visual spatial encodings of the masked target $f^{\tilde{v}}_{m}(\gamma \odot v)$ can be correlated against the radio spatial encodings covering the entire sensing scene $f^{r}_n(r)$ in order to produce an attention map (with appropriate padding)
\vspace{-0.20cm}
\begin{align}
  h_n(r, v) &= \operatorname{conv2d} \big( f^{r}_n(r), f^{\tilde{v}}_{m}(\gamma \odot v) \big), \hspace{0.25cm} n \in \Omega, m \in \tilde{\Omega}
  \label{eq:2d_attention}
\end{align}

\vspace{-0.2cm}
To measure best cross-modal regional agreement, the attention map is maximised over spatial bins
\vspace{-0.20cm}
\begin{align}
  S(r, v) = \underset{n \in \Omega}{\max} \; h_n(r, v)
  \label{eq:max_2d_attention}
\end{align}

\vspace{-0.50cm}
\subsection{Target self-estimation}

\vspace{-0.1cm}
Once the backbone networks are learnt and their spatial features are stable, we can use cross-modal attention maximisation (cf., Eqs.~\ref{eq:2d_attention}~\&~\ref{eq:max_2d_attention}) to self-generate target coordinate estimates.
This self-labelling is inherently noisy, but remarkably powerful.
Particularly, a downstream localiser network is able to smooth these self-estimates when trained over a sufficiently large number of data points---as determined by the mutual information with perfect coordinates~\cite{guan2018said}.

\noindent \textbf{Rescaling and calibration.} Target coordinate estimates are obtained in the spatial feature grid $h \times w$.
We rescale to bring back to original grid $H \times W$, and perform one-off calibration for systematic offsets on entire dataset. 

\vspace{-0.15cm}
\subsection{Localiser network} \label{sec:loc_net}

\vspace{-0.15cm}
We construct the dataset $(r, \hat{y}) \in \mathcal{D}_{\text{loc}}$ from tuples of radio heatmaps $r$ and their target self-labels $\hat{y}$.
The localiser network is trained to regress $\hat{y}$ from $r$ using a mean squared error (MSE) loss.

\vspace{-0.20cm}
\section{Benchmarks} \label{sec:experiments}

We discuss our baselines and empirical evaluation.
We refer the reader to Appendix~\ref{sec:implement_details} for implementation details.

\vspace{-0.1cm}
\subsection{Baselines} \label{sec:baselines}

Radar target detection is a historic and thoroughly investigated topic as it pertains to many civil and military applications.
The objective is to predict a target's position and velocity. 
However, extracting wanted information (i.e., the target) from unwanted information (i.e., clutter) is a challenging task.
Due to radio propagation phenomena, both could exhibit comparable statistical behaviour.
We implement expert statistical techniques used by various industries in millions of products, and designate as our first strong standard baseline.
Equally, a fully-supervised localiser network trained on groundtruth coordinates naturally forms our second deep learning-based baseline.
We also adapt to our dataset a third radio-visual fusion scheme called RODNet~\cite{wang2021rodnet}.
In what follows, we describe these approaches. 

\noindent \textbf{Statistical.} 
Extracting information from a radio response representation is a multi-step procedure. 
First, radio targets in two different domains, range-angle and range-velocity, are binarised via a threshold technique (e.g., CFAR~\cite{rohling1983radar}) and then clustered (e.g., via DBSCAN~\cite{ester1996density}) to form one point cloud per target.  
Targets are then matched between the two different domains over the same and hopefully unique range. 
Point cloud centroids are used to track targets. 

Considering such multi-step procedure, the following shortcomings come to mind. First, how should we detect the wanted target from the matched targets (e.g., how to remove clutter). 
Second, some information is ignored when assigning a target centroid (e.g., information from the shape of the point clouds).
Third, setting the optimal thresholds, guard bands, training bands, number of points per cluster~\cite{rohling1983radar} is an exceedingly brittle exercise.
It is our hope that end-to-end learning is able to address some of these shortcomings.

For the statistical baseline to become more competitive with learning-based approaches, we make it ``Genie-aided'', i.e., the peak closest to groundtruth is assigned as a target.
Genie-aided algorithms are common practice in information theory literature to study upper performance bounds~\cite{devroye2006achievable}.

\noindent \textbf{Supervised.} 
In radio sensing, labelling empirical heatmaps (e.g., object type, centre, bounding box) is infeasible at scale as we cannot interpret the scene by manual inspection.  
However, we consider the supervised network as a useful upper bound on the performance of self-supervised localisation.

Compared to computer vision, radio imaging has no prescribed or de facto neural architectures to use for evaluation.
We therefore use Microsoft's AutoML tool NNI (Neural Network Intelligence)~\cite{nni2021} to search for strong candidate architectures. 
Specifically, we searched for optimizers, loss functions, learning rates, momentums, neural architectures via resolution branching, and activation functions. 
The performance of the supervised baseline in Sec.~\ref{sec:results} corroborates the quality of the search.
Detailed description of the architectural search space is given in Appendix~\ref{sec:nni_appendix}.

\noindent \textbf{Radio-visual fusion.} 
RODNet uses a student-teacher network configuration~\cite{wang2021rodnet}. 
The teacher combines object detection in vision and statistical peak detection in radio to derive object class and location estimates.
The radio-only student network is trained on the teacher's estimates. 
We use the student network as a baseline and characterise against pseudo groundtruth labels.
Like our system, our RODNet implementation operates on a single heatmap snapshot without spatio-temporal convolution.
Compared to statistical CFAR baseline---using a genie-aided peak selection where the peak closest to the target is always assigned---RODNet's vision+radio teacher implements a peak fusion to approximate the optimal joint camera-radio detector.

\vspace{-0.15cm}
\subsection{Empirical data}

\vspace{-0.15cm}
For further empirical validation of our radio-visual SSL algorithm, we use the parking lot scenario of the Camera-Radar of the University of Washington (CRUW) dataset~\cite{wang2021rodnet}.
Tab.~\ref{tab:maxray_vs_cruw} in Appendix~\ref{sec:cruw} compares CRUW to~\dataset.

\noindent \textbf{Pseudolabels construction.}
Since empirical data does not come with groundtruth correspondence labels, we employ the following pseudolabelling procedure (based on RODNet).
First, we detect and segment objects from images using a Mask R-CNN object detector~\cite{he2017mask}.
Second, we detect radar peaks using CFAR and cluster them into groups using DBSCAN.
Third, we perform intrinsic camera calibration to convert camera objects into x-y coordinates.
Fourth, we match image segmentations to radar peak clusters. 

\noindent \textbf{Pseudosupervision.}
As a replacement for the supervised baseline on~\dataset, we use the matched range-angle pseudolabels from the above procedure to train a pseudosupervised localiser net that takes as input CRUW radar heatmaps.

\vspace{-0.15cm}
\subsection{Metrics}

\vspace{-0.15cm}
We measure location estimation performance using two error statistics: $50$th and $90$th percentiles (abbrv. \%ile), and on the validation set unless otherwise stated.
Formally, let $X_e$ denote the location error as a random variable, $x_e$ denote the \%ile error, $\Pr$ the error probability distribution, $F_{X_e}$ its cumulative probability distribution, and $p_e \in \{0.5, 0.9\}$ a probability value it assumes.
Then $p_e \coloneqq F_{X_e}(x_e) = \Pr(X_e \le x_e)$.
Throughout evaluation, we will simply quote these error statistics ($x_e \big|_{p_e \in \{0.5, 0.9\}}$) as the $50$th and $90$th $\%$ile errors.

\vspace{-0.1cm}
\section{Results} \label{sec:results}

\vspace{-0.1cm}
Having described our benchmarking setup in Sec.~\ref{sec:experiments}, we turn next to discussing results.
For~\dataset~and CRUW, results are computed on 6k and 1.8k data points of validation sets, respectively.

\vspace{-0.15cm}
\subsection{Localisation performance}

\vspace{-0.15cm}
We examine the overall performance of our MCL-based self-labelled localiser net and compare it against: fully supervised, RODNet, and statistical baselines. 
The self-labelled net, supervised, and RODNet share identical downstream architecture and training configurations.
This common architecture is only specialised with different convolutional kernel sizes across datasets due to differences in radio configurations (cf., \dataset~vs.~CRUW in Appendix~\ref{sec:cruw}).
We denote the statistical baseline by Constant False Alarm Rate (CFAR).
Tab.~\ref{tab:loc_perf_summary} summarises the performance in terms of $50$th $\%$ile and $90$th $\%$ile localisation errors on the validation sets.
\vspace{-0.0cm}
\begin{table}[t]
\begin{minipage}[b]{.45\textwidth}
    \centering
    \caption{\small Performance summary on~\dataset~and CRUW.\\ MCL sets a new SOTA perf. for \emph{label free} localisation.}
    \medskip
    \vspace{-0.50cm}
  \scriptsize
  \setlength{\tabcolsep}{2pt}
  \begin{tabular}{lcHHrr}
    \toprule
    \;                           &           & \multicolumn{2}{B}{\dataset~perf. (error in m)}              & \multicolumn{2}{c}{CRUW perf. (error in m)}                   \\
    \cmidrule(lr){3-4} \cmidrule(lr){5-6}
    \multicolumn{1}{c}{Method}   &Label free & $50$th $\%$ile               & $90$th $\%$ile                & $50$th $\%$ile   & $90$th $\%$ile                             \\ 
    \midrule
    Supervised$^{\ast}$             & \xmark    & 0.289 {\tiny \textpm 0.017}  & 0.922 {\tiny \textpm 0.042}   & 1.382 {\tiny \textpm  0.128}   &  5.402 {\tiny \textpm 0.063} \\
    \rowcolor{gray!30!white}
    MCL                          & \cmark    & \textbf{0.942} {\tiny \textpm 0.016}  & \textbf{4.681} {\tiny \textpm 0.158}   & \textbf{2.558} {\tiny \textpm  0.072}   & 10.969 {\tiny \textpm 0.032} \\
    CFAR$^{\dagger}$             & \cmark    & 2.709 \phantom{{\tiny \textpm 0.000}}    & 8.062 \phantom{{\tiny \textpm 0.000}}     & 4.659 \phantom{{\tiny \textpm 0.000}}      & 6.161 \phantom{{\tiny \textpm 0.000}} \\
    RODNet                       & \cmark    & 3.012 {\tiny \textpm 0.014}  & 8.913 {\tiny \textpm 0.107}   & 3.281 {\tiny \textpm  0.334}   &  7.791 {\tiny \textpm 0.355} \\
    \bottomrule
    \label{tab:loc_perf_summary}
  \end{tabular}
  \scriptsize{
  \vspace{-0.35cm}\\
  \parbox{1\textwidth}{$^{\ast}$pseudosupervised in CRUW \hspace{0.15cm} $^{\dagger}$Genie-aided}
  }
\end{minipage}
\begin{minipage}[t]{.225\textwidth}
\vspace{-0.45cm}
    \centering
    \caption{\small Backbone training configurations: MCL, SCL, CL}
    \medskip
    \vspace{-0.50cm}
  \scriptsize
  \setlength{\tabcolsep}{2pt}
  \begin{tabular}{lrr}
    \toprule
      & \multicolumn{2}{c}{\dataset~perf. (error in m)} \\ %
    \cmidrule(lr){2-3}  
    \multicolumn{1}{l}{Backbone}       & $50$th $\%$ile                    & $90$th $\%$ile     \\
    \midrule
    MCL                                & 0.942 {\tiny \textpm 0.016}   &  4.681 {\tiny \textpm 0.158}  \\
    SCL                                & 1.571 {\tiny \textpm 0.050}   &  3.539 {\tiny \textpm 0.062}  \\ 
    CL                                 & 3.111 {\tiny \textpm 0.358}   & 17.498 {\tiny \textpm 0.317}  \\  
    \bottomrule
    \label{tab:loc_perf_backbone_configs}
  \end{tabular}
\end{minipage}\hfill
\begin{minipage}[t]{.235\textwidth}
\vspace{-0.45cm}
    \centering
    \caption{\small Backbones with Yolov5 bounding boxes.}
    \medskip
    \vspace{-0.50cm}
  \scriptsize
  \setlength{\tabcolsep}{2pt}
  \begin{tabular}{lrr}
    \toprule
     & \multicolumn{2}{c}{\dataset~perf. (error in m)} \\ %
    \cmidrule(lr){2-3}  
    \multicolumn{1}{l}{Backbone}   & $50$th $\%$ile   & $90$th $\%$ile    \\
    \midrule
    MCL$^{Y}$                      & 1.351            & 7.649             \\
    SCL$^{Y}$                      & 2.188            & 5.462             \\ 
    \bottomrule
    \label{tab:loc_perf_backbone_yolov5}
  \end{tabular}
  \scriptsize{
  \vspace{-0.30cm}\\
  \parbox{1\textwidth}{$^{Y}$using Yolov5 bounding boxes}
  }
\end{minipage}
\end{table}
Not surprisingly, the fully-supervised net performs most favourably with around 30cm and 1.4m median errors, respectively on~\dataset~and~CRUW.
Note the drop in supervised performance between~\dataset~and CRUW is largely due to the halved angular resolution of CRUW (cf., Appendices~\ref{sec:radio-visual_analytic_relationship}~\&~\ref{sec:cruw}).
MCL comes second with approx. 0.94m and 2.5m median errors, respectively on ~\dataset~and CRUW.
This is remarkable given that MCL has automatically learned how to localise targets by simply observing paired radio-visual data.
Genie-aided CFAR performs worse with roughly 2.8$\times$ and 1.8$\times$ MCL's median errors, respectively on ~\dataset~and CRUW.
RODNet is also worse with 3.2$\times$ and 1.3$\times$ MCL's median errors, respectively on ~\dataset~and CRUW.
As discussed in Sec.~\ref{sec:baselines}, RODNet teacher employs a conventional radio-visual fusion scheme that relies on radio CFAR detection aided by vision.
On the higher angular resolution of \dataset, RODNet's fusion scheme seems to be far less effective than our joint embedding architecture.

\subsection{Ablations and analysis} \label{sec:ablations_analysis}

\vspace{-0.15cm}
We now conduct experiments to better understand MCL's performance against alternatives, its dependence on masking accuracy and self-labelling density, its modelling capacity, and its sensitivity to radio-visual commonalities.

\vspace{-0.0cm}
\begin{figure*}[t]
\begin{minipage}[t]{.24\textwidth}
  \centering
  \includegraphics{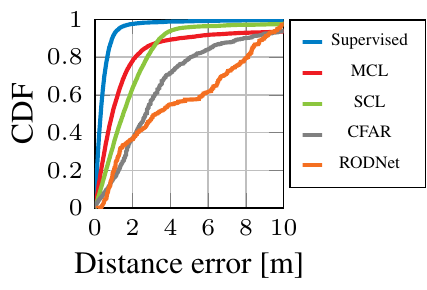}
  \vspace{-0.65cm}
  \caption{\small Localisation error CDFs of methods: Supervised, MCL, SCL, CFAR, and RODNet.}
  \label{fig:perf_cdfs}
\end{minipage}\hfill
\begin{minipage}[t]{.36\textwidth}
  \centering
  \includegraphics{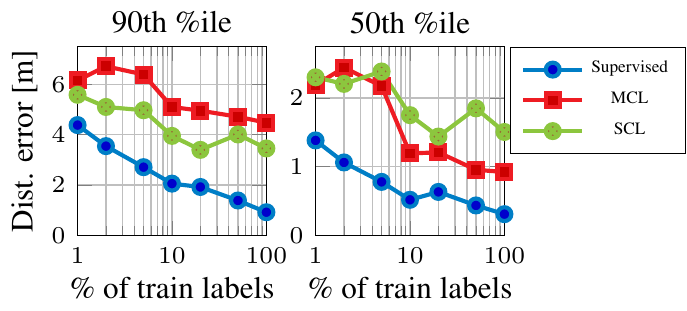}
  \vspace{-0.65cm}
  \caption{\small Effect of number of training labels on localisation performance for Supervised, MCL, and SCL.}
  \label{fig:label_density}
\end{minipage}\hfill
\begin{minipage}[t]{.34\textwidth}
  \centering
  \includegraphics{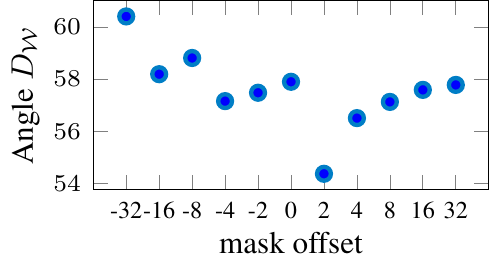}
  \vspace{-0.25cm}
  \caption{\small Effect of radio-visual mutual information on self-labelling for MCL, on \dataset's validation set and as measured by $D_\mathcal{W}$.}
  \label{fig:mi_uShaped}
\end{minipage}
\end{figure*}

\noindent \textbf{Masked contrast vs. other contrastive learning flavours.}
We have found MCL to be an effective radio-visual learning strategy on synthetic and noisier empirical data.
We would like to understand, however, how MCL compares to other forms of contrastive learning from the literature.
To this end, we first consider spatial contrastive learning (SCL) that has appeared in multiple recent works that use 2-D backbone modelling~\cite{afouras2020self,afouras2021self,windsor2021self}.
SCL performs contrastive learning in 2-D to incentivise cross-modal spatial attention.
We adapt SCL to the radio-visual problem setting and provide formal definition and illustrative comparisons in Appendix~\ref{sec:cl_background}.
We also consider vanilla contrastive learning (CL) without masking~\cite{chen2020simple}.
I.e., the following investigates the performance of model variants: SCL and CL.
We ask: \emph{What role does masking play during contrastive radio-visual learning?}

Tab.~\ref{tab:loc_perf_backbone_configs} analyses the performance of the three backbone configurations on~\dataset.
We note that vanilla CL performs poorly with 3.1m median error.
We attribute the high localisation error to the lack of target sensitivity of CL during training.
MCL, on the other hand, is trained to attend to targets through masking and exhibits a $50$th $\%$ile error of 0.94m, interestingly 1/3rd better than SCL at 1.57m.
A closer look at MCL's $90$th $\%$ile error at 4.6m reveals that it is also around 1.3$\times$ ``lazier'' than SCL at tracking higher $\%$ile targets.
However, we note that SCL has \emph{failed} to train on the noisier empirical CRUW dataset, while MCL has given a new SOTA $50$th $\%$ile performance as shown in Tab.~\ref{tab:loc_perf_summary} (despite CRUW's low angular resolution).
We conjecture that MCL's 2-layer MLP projector supports denoising---a crucial feature the more computationally efficient SCL lacks.
Classically, noisy radar data could have many spurious ghost targets~\cite{rohling1983radar}. 
Hence, a parameter-heavy projector head may prove necessary to stabilise learning. 
Fig~\ref{fig:perf_cdfs} depicts the error cumulative density functions (CDFs) of all methods.

\noindent \textbf{Impact of noisy masks.}
Taking advantage of the controllability of~\dataset, we have so far utilised perfect masks generated during synthesis.
We now investigate the impact of using noisy mask estimates during contrastive learning.
To this end, Tab.~\ref{tab:loc_perf_backbone_yolov5} lists the localisation performance of MCL and SCL using masks from Yolov5, similar to how we integrated CRUW into our SSL pipeline (see Appendices~G~\&~J).
We observe around 0.4m and 0.6m degradation in the median localisation performance, respectively for MCL and SCL.

\noindent \textbf{Impact of label density.}
We investigate performance enhancements as a function of increased number of noisy labels.
Using~\dataset's 24k training set, we sweep the amount of labels and self-labels used to train the localiser nets of supervised, MCL, and SCL.
Then we evaluate on the validation set to gauge the localisation performance sensitivity to the amount of available training (self-)labels.
We cover the training points in logarithmic steps.

\begin{figure*}[t]
\begin{minipage}{.525\linewidth}
\begin{center}
  \captionsetup{type=table}
  \caption{\small Quantifying how far self-labels deviate from groundtruth.}
  \vspace{-0.35cm}
  \scriptsize
  \setlength{\tabcolsep}{2pt}
  \begin{tabular}{lBBBBBBCCCCCC}
    \toprule
    \multicolumn{1}{c}{Config.}   & \multicolumn{6}{B}{Training}                & \multicolumn{6}{C}{Validation}             \\[-0.0cm]
    \cmidrule(lr){2-7} \cmidrule(lr){8-13} 
                    & \multicolumn{3}{B}{Range}     & \multicolumn{3}{B}{Angle}     & \multicolumn{3}{C}{Range}     & \multicolumn{3}{C}{Angle}   \\[-0.0cm]
    \cmidrule(lr){2-4} \cmidrule(lr){5-7} \cmidrule(lr){8-10} \cmidrule(lr){11-13}
                    & $D_\mathcal{W}^\downarrow$ & $D_\text{KL}^\downarrow$ & $\text{MI}^\uparrow$    & $D_\mathcal{W}^\downarrow$ & $D_\text{KL}^\downarrow$ & $\text{MI}^\uparrow$    & $D_\mathcal{W}^\downarrow$ & $D_\text{KL}^\downarrow$ & $\text{MI}^\uparrow$    & $D_\mathcal{W}^\downarrow$ & $D_\text{KL}^\downarrow$ & $\text{MI}^\uparrow$   \\
    \midrule
    MCL             & 28.509 & 7.294 & 0.947    & 59.359 & 5.572 & 1.120        & 27.940 & 5.319 & 0.931    & 57.901 & 3.571 &  1.121 \\
    SCL             & 19.063 & 7.281 & 1.217     & 40.115 & 5.548 & 1.567        & 18.691 & 5.282 & 1.226    & 39.387 & 3.541 &  1.528 \\ 
    \midrule
    MCL$^{Y}$      & 34.930  & 7.287  & 0.934    & 63.613 & 5.559 & 0.985   & 34.814 & 5.303 & 0.913  & 62.095 & 3.572  & 0.987  \\
    SCL$^{Y}$      & 20.867  & 7.273  & 1.117    & 49.472 & 5.566 & 1.512   & 21.035 & 5.289 & 1.124  & 48.213 & 3.562  & 1.476  \\ 
    \bottomrule
    \label{tab:distribution_deviation_metrics}
  \end{tabular}
  \scriptsize{
  \vspace{-0.35cm}\\
  \parbox{1\textwidth}{$^{Y}$using pseudo bounding boxes obtained from Yolov5} 
  }
\end{center}
\end{minipage}\hfill
\begin{minipage}[t]{.425\linewidth}
    \centering
    \includegraphics{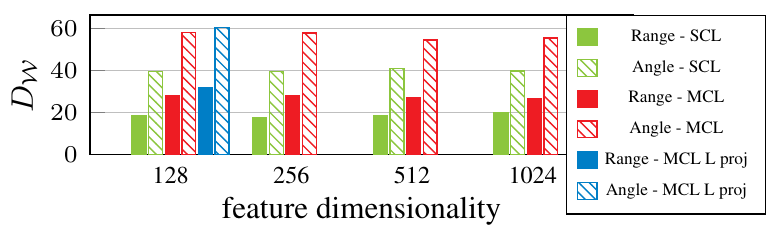}
    \vspace{-0.75cm}
    \caption{\small Effect of dimensionality on self-labelling for MCL \& SCL, on \dataset's validation set and as measured by $D_\mathcal{W}$. L proj in legend denotes 4$\times$ projector head.}
    \label{fig:feature_dim}
\end{minipage}
\end{figure*}

With the noisy labels of MCL and SCL, it can be shown that the localiser nets learn to compensate for such noise by using sufficiently large number of data points~\cite{guan2018said}.
The number of required data points is a function of the mutual information between noisy target coordinates and perfect coordinates~\cite{guan2018said}.
Fig.~\ref{fig:label_density} examines this effect for supervised as a reference baseline, and MCL and SCL.
We observe that MCL has a better label density tolerance than SCL w.r.t. the $50$th $\%$ile performance.
The opposite holds true, i.e., SCL is better than MCL w.r.t. the $90$th $\%$ile performance.
This finding mirrors the localisation error analysis of~Fig.~\ref{fig:perf_cdfs}.
On~\dataset, MCL seems to be a better self-localiser of the bulk of the distribution of the validation set, while SCL seems to cope better with corner cases.

\noindent \textbf{Self-labels deviation from groundtruth.}
We have uncovered qualitative differences between the ability of MCL and SCL to self-localise targets using cross-modal attention.
We turn next to quantify how far the self-labels of MCL and SCL deviate from groundtruth labels.
The following analysis sheds further light on the performance differences between masked projector contrast and spatial contrast.

We analyse the deviation of self-labels from groundtruth labels through the lens of three metrics.
Let $p_\text{gt}(y)$ and $p_\text{est}(\hat{y})$ denote the distributions of groundtruth labels $y$ and self-label estimates $\hat{y}$, respectively.
The shift between $p_\text{gt}$ and $p_\text{est}$ can be quantified using the 1-D Wasserstein distance 
$D_{\mathcal{W}}(p_\text{gt}, p_\text{est}) = \int_0^1 c \big( \, | F_{\text{gt}}^{-1}(x) - F_{\text{est}}^{-1}(x) | \, \big) dx$,
where $F$ is the CDF function, and $c$ denotes a cost function---we use a quadratic cost below.
We employ $D_{\mathcal{W}}$ because of its robustness and specificity on empirical measurements~\cite{Peyre20_ComputationalOT}.
We also use two more conventional information-theoretic measures: the Kullback-Leibler (KL) divergence $D_\text{KL}( p_\text{gt}(y) || p_\text{est}(\hat{y}) )$ and mutual information (MI) $I(y; \hat{y}) = D_\text{KL}(p_{(\text{gt}, \, \text{est})} || p_\text{gt} \, p_\text{est})$~\cite{thomas2006elements}.
KL quantifies the shift between $p_\text{gt}$ and $p_\text{est}$, similar to the Wasserstein distance.
MI measures how dependent $p_\text{est}$ is on $p_\text{gt}$ (in nats below).

Tab.~\ref{tab:distribution_deviation_metrics} evaluates numerically the three distribution deviation metrics.
The metrics are computed for the training and validation sets separately.
The metrics are also computed for range and angle coordinates separately.
On~\dataset, we can see that SCL outperforms MCL consistently across metrics, sets, and coordinates.
Specifically, both $D_\mathcal{W}$ and $D_\text{KL}$ are lower for SCL, indicating better match to groundtruth.
Similarly, MI is higher for SCL, indicating better match to groundtruth.
Similar observations hold for MCL$^{Y}$ and SCL$^{Y}$ using mask estimates from Yolov5.
For qualitative comparison of distributions, consult the empirical histograms in Fig.~\ref{fig:selfLabels_vs_groundtruth_distributions} in Appendix~\ref{sec:appendix_selfLabel}.

\noindent \textbf{Impact of dimensionality.}
We investigate if SSL backbone capacity limits performance.
To do so, we train backbone configurations and measure their self-label deviation from groundtruth as a function of: (a) feature dimensionality per spatial bin (denoted by $C$ in Sec.~\ref{sec:representation_learning}) for MCL and SCL, and (b) the dimensionality of the 2-layer MLP projector head for MCL.  
Fig.~\ref{fig:feature_dim} depicts the Wasserstein distances across a number of (a)~\&~(b) configurations.
For SCL, doubling $C$ up to 1024 features per spatial bin has negligible effect on range and angle self-label distances to groundtruth.
For MCL, there is a mild reduction in distances as a function of feature dimensionality, and no effect for using a larger projector head.
We, therefore, conclude that the performance of self-labels is fundamentally limited by the underlying resolution of radio imaging rather than the model's learning capacity.

\noindent \textbf{Impact of cross-modal commonalities.}
We investigate the relationship between cross-modal commonalities and localisation performance.
The InfoMin principle tells us how to ``regularise'' contrastive learning in order to obtain optimal downstream performance~\cite{tian2020makes}.\footnote{building on earlier information bottleneck literature~\cite{tishby2000information,alemi2016deep,fischer2020conditional}}
According to InfoMin, there are three regimes of MI captured during learning: (1) missing info, (2) sweet spot, and (3) excessive info~\cite{tian2020makes}.
These three regimes can be empirically observed as a U-shaped curve for a given downstream task.
We control the amount of radio-visual MI through masking in the vision domain.
Specifically, we train MCL model variants with the groundtruth target masks progressively enlarged or shrunk, i.e., by positively or negatively padding the masks.
We then obtain self-labels for these MCL variants and measure their $D_\mathcal{W}$ as before.
Fig.~\ref{fig:mi_uShaped} depicts $D_\mathcal{W}$ of the angle distribution as a function of target mask offsets.
We observe a U-shaped curve whose minima is at an offset of 2 pixels.
This corroborates that masking in vision enhances target sensitivity (ablated in Tab.~\ref{tab:loc_perf_backbone_configs}), and further illustrates the degradation as we increase ($+$ offsets) or reduce ($-$ offsets) radio-visual MI.

\vspace{-0.25cm}
\section{Discussion}

\vspace{-0.2cm}
\noindent \textbf{6G sensing.}
\hspace{-0.125cm}
Making cellular basestations ``see'' the surrounding environment while sending data is a major feature in 6G networks.
There are non-trivial protocol-level challenges in 6G network design in order to support sensing (see Appendix~E).
In this paper, we concentrate on the higher-level challenge of \emph{automatically} building target localiser models using radio heatmaps that are accompanied by visual images, i.e., paired radio-visual data collected at a basestation equipped with a camera.
Through cross-modal attention, we show how to estimate self-labels for training a downstream radio localiser network.
Specifically, we demonstrate that the performance of the localiser network is \emph{not upper bounded} by the accuracy of self-labels, and that using larger number of noisy self-labels enhances estimation.
This finding is in line with prior work~\cite{guan2018said}, and serves to reaffirm the paradigm of self-supervised radio-visual learning for scalable radio sensing.
Our synthetic radio-visual dataset helps establish the performance trends of radio-visual SSL localisation by virtue of a controlled groundtruth. 
That is, out dataset is dedicated to the study and refinement of radio-visual SSL algorithms, and \emph{not} to the production of 6G perception models.
Our SSL algorithm, on the other hand, is readily applicable to empirical data with no groundtruth (cf., Tab.~\ref{tab:loc_perf_summary}).
We believe that our radio-visual SSL objective provides a viable route towards vast data scalability for 6G sensing.

\noindent \textbf{Limitations.}
We note that radio sensing capabilities are fundamentally set by the choice of configurations in Tab.~\ref{tab:radio_params}.
We have opted to base this somewhat conservative choice on 5G Advanced specifications~\cite{kim2019new} in order to inform cellular stakeholder discussions.
We would, however, note that much improved radio sensing performance can be attained through increased bandwidth and/or denser antenna arrays, such as in Terahertz or even higher Millimetre-wave bands~\cite{elkhouly2022fully,shahramian2019fully}.
We would refine our dataset and results in light of future consensus on 6G sensing specifications.

\noindent \textbf{Broader impact.}
Our work has a broader societal impact in that it has the potential to alleviate some of risks associated with the surveillance economy.
Specifically, once trained and deployed, our radio sensing system offers a scalable alternative to pervasive vision surveillance that is inherently privacy-preserving,
while achieving many of the sought-after safety and security benefits. 

\vspace{-0.20cm}
\section{Conclusion}

\vspace{-0.2cm}
In this paper, we present a new radio-visual learning task for emerging 6G cellular networks.
The task tackles the problem of accurate target localisation in radio, employing a novel learning paradigm that works by simply ingesting large quantities of paired radio-visual data.
This is in stark difference to supervised and/or classic statistical methods whose success hinges on laborious labelling and/or modelling of empirical measurements, which are expensive to scale.
We demonstrate strong label-free target localisation performance on synthetic and empirical data.
Our novel target localisation paradigm is made possible by a new dataset and benchmark intended to foster future research on radio sensing for next generation cellular systems.

\section*{Acknowledgments}

We thank Dmitry Chizhik for his analytic modelling of the relationship between radio imaging and vision in Appendix~\ref{sec:radio-visual_analytic_relationship}.  
We would like to thank Howard Huang, Akash Singh and Prof. Mani Srivastava for their helpful discussions.

{\small
\bibliographystyle{ieee_fullname}
\bibliography{references}
}

\begin{appendices}

This supplementary material consists of 11 appendices.
It provides expanded discussion, background details, results, illustrations, and documentation for radio-visual data and algorithms.

\section{Radio-visual analytic relationship} \label{sec:radio-visual_analytic_relationship}

\begin{figure}[b]
\begin{minipage}{\textwidth}
  \centering
  \captionsetup{justification=centering}
  \vspace{-0.20cm}
  \subfloat[\footnotesize grey-scale camera image]{
    \includegraphics[width=0.31\textwidth]{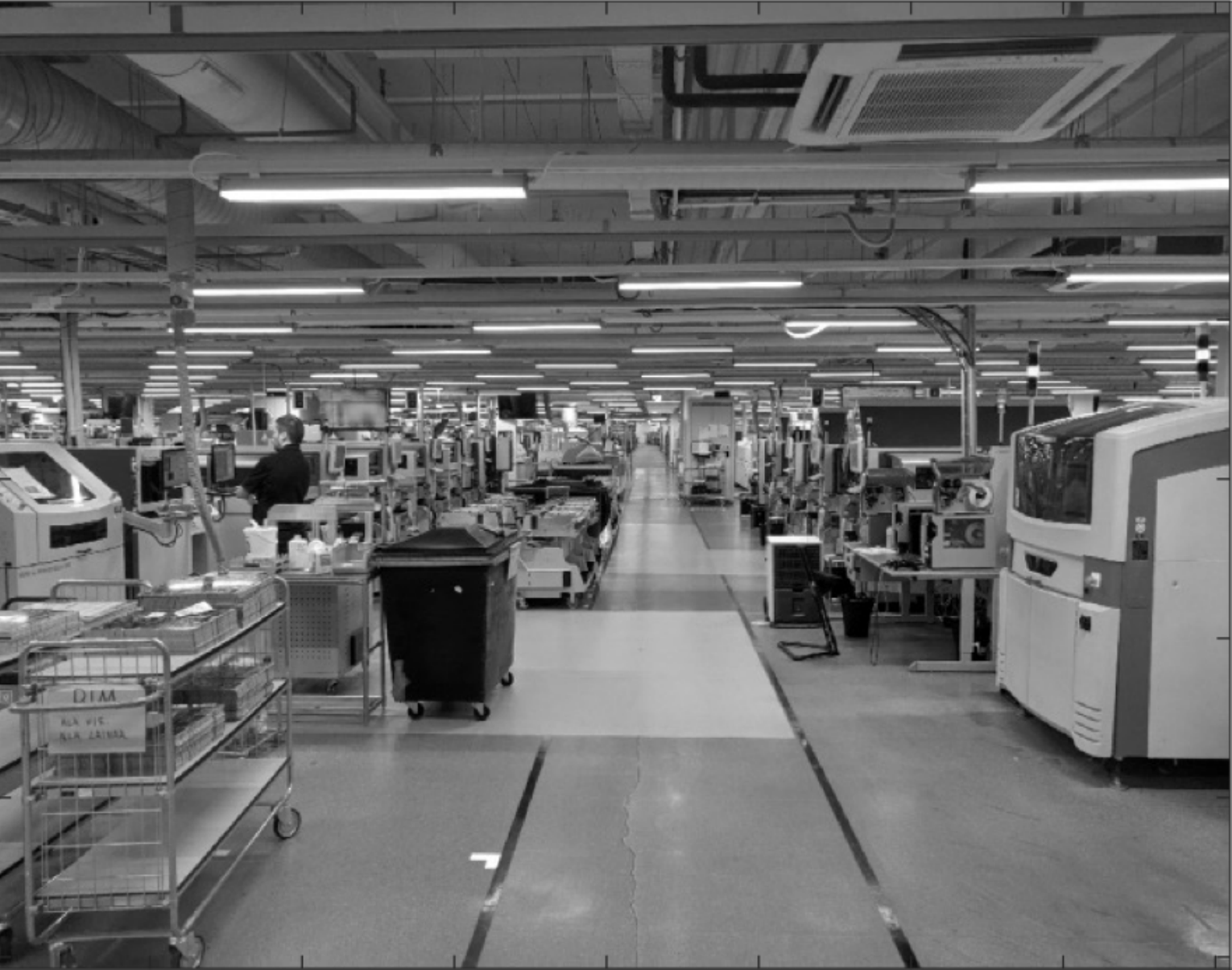}
    \label{fig:factory}
  }
  \hspace{-0.20cm}
  \subfloat[\footnotesize moderate blur]{
    \includegraphics[width=0.31\textwidth]{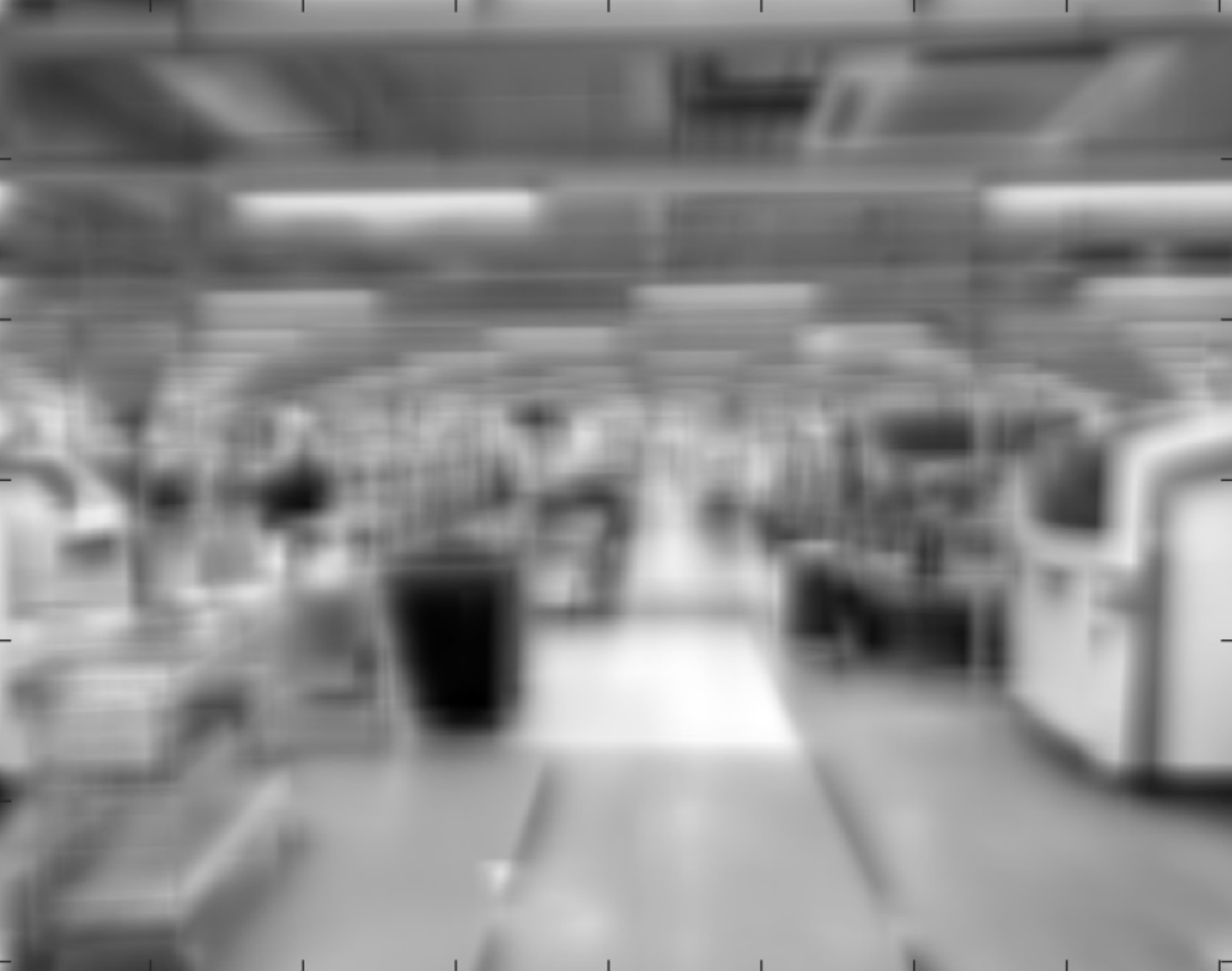}
    \label{fig:factory_blur0}
  }
  \hspace{-0.20cm}
  \subfloat[\footnotesize high blur]{
    \includegraphics[width=0.31\textwidth]{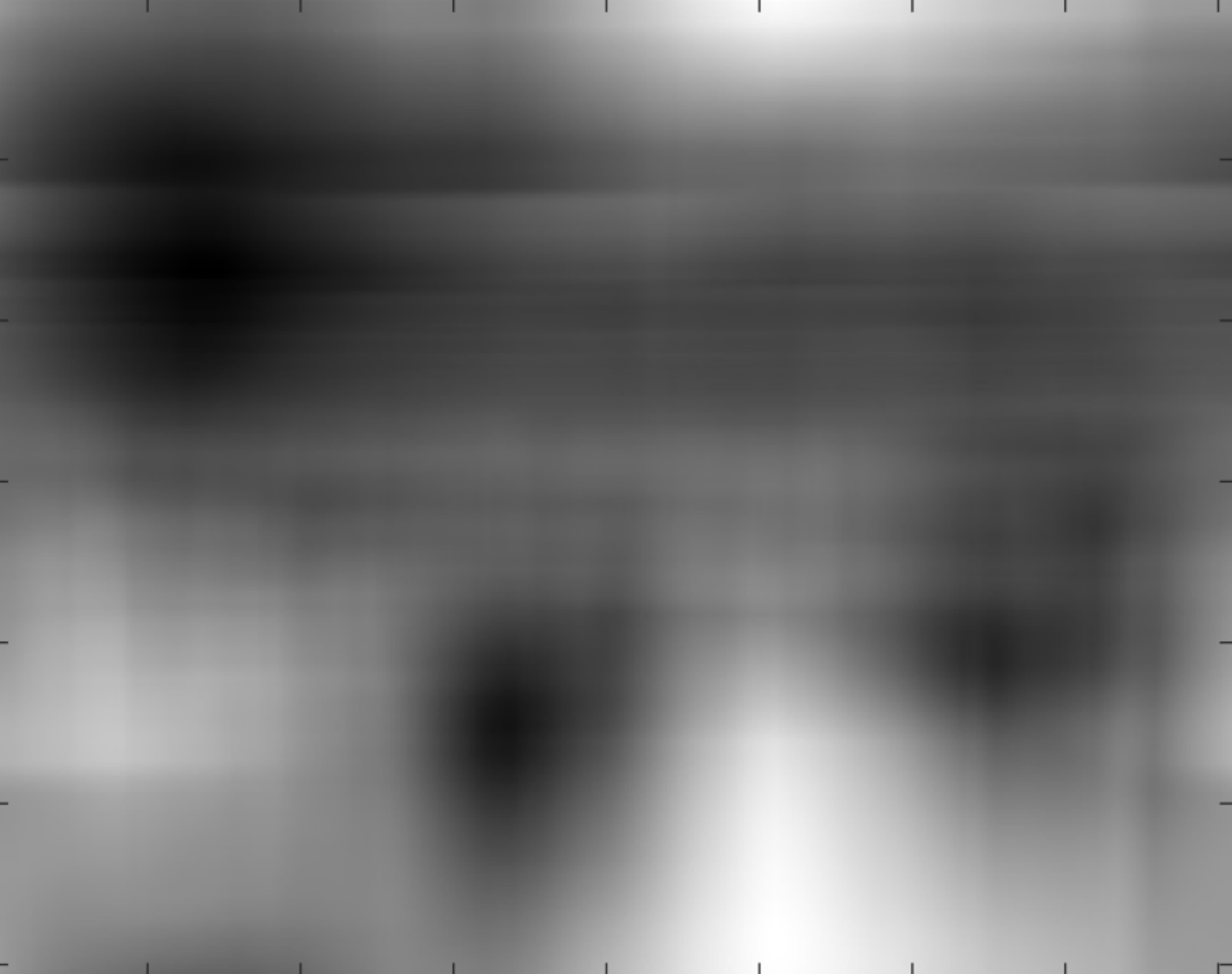}
    \label{fig:factory_blur1}
  }
  \vspace{-0.0cm}
  \caption{\raggedright Radio-visual relationship. A grey-scale camera image~\protect\subref{fig:factory} undergoes blurring in~\protect\subref{fig:factory_blur0}~\&~\protect\subref{fig:factory_blur1} to simulate the effect of RF's limited angular resolution when using radio to image the environment. \protect\subref{fig:factory_blur0}~shows moderate blur while~\protect\subref{fig:factory_blur1} shows significant blur as a result of angular resolutions $\Delta \phi=1^{\circ}$ and $\Delta \phi=10^{\circ}$, respectively.}
  \label{fig:radio-visual_relationship}
\end{minipage}
\end{figure}

In radio imaging, there are two main phenomena that govern our ability to resolve objects in space.
First, range resolution $\Delta r$ is determined by bandwidth $B$ and obeys $\Delta r = c/2B$, where $c$ is the speed of light.
In typical millimetre-wave frequencies for 6G, $B \approx 1\text{GHz}$ which gives $\sim0.3$ metre resolution.
Second, the angular resolution $\Delta \phi$ is considerably worse and is generally related to our ability to pack antennae in a reasonable form factor.
That is, the imaging performance disparity between vision and radio is largely a function of disparities in angular resolution.
To see this, let $I(x, y)$ be an image of a sensing scene, where $x$ and $y$ are its horizontal and vertical dimensions, respectively.
Let $w$ be the so-called beamwidth of an RF horn antenna.
Then the antenna response $h(x,y) = e^{-(x^2+y^2)/(2w^2)}$ is a ``distortion'' function associated with RF's resolution-limited imaging of a given scene.
Specifically, $h(x,y)$ will act as a blurring function that convolves with the original image according to 
\begin{align}
  I^{\prime}(x, y) = I(x, y) * h(x,y) 
  \label{eq:rf_blurring-function}
\end{align}
where $I^{\prime}$ is the degraded image and $*$ is the convolution operator.

Fig.~\ref{fig:radio-visual_relationship} contrasts normal camera imaging against RF's resolution-limited imaging.
Left-most Fig.~\ref{fig:factory} shows a grey-scale image of a factory.
Assuming 1 degree angular resolution ($\Delta \phi=1^{\circ}$), Fig.~\ref{fig:factory_blur0} in the middle illustrates the blurring effect of Eq.~\ref{eq:rf_blurring-function} on the camera image.
Under higher angular resolution distortion $\Delta \phi=10^{\circ}$, the right-most Fig.~\ref{fig:factory_blur1} shows significant blurring as a result of a coarser beamwidth $w$ acting on $I$.

\newpage

\section{Radio-visual subspace analysis} \label{sec:appendix_subspace}

\begin{figure}[b]
\begin{minipage}{\textwidth}
  \centering
  \captionsetup{justification=centering}
  \subfloat[\footnotesize channel subspace - radio]{
    \includegraphics[width=0.4\textwidth]{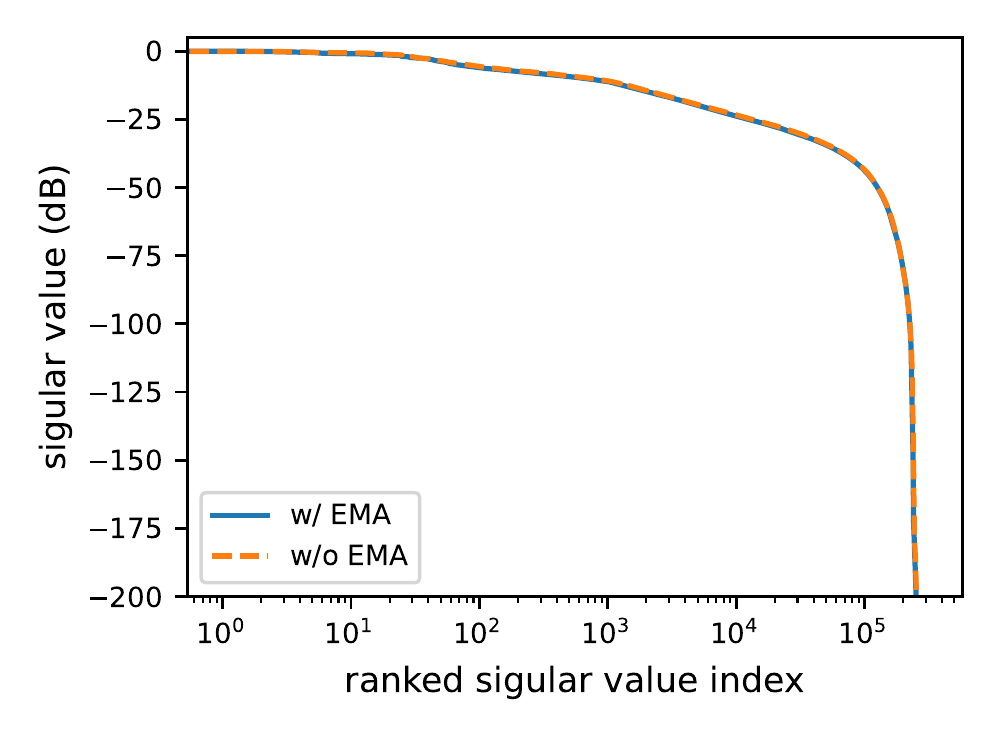}
    \label{fig:subspace_channel_r}
  }
  \hspace{-0.25cm}
  \subfloat[\footnotesize channel subspace - visual]{
    \includegraphics[width=0.4\textwidth]{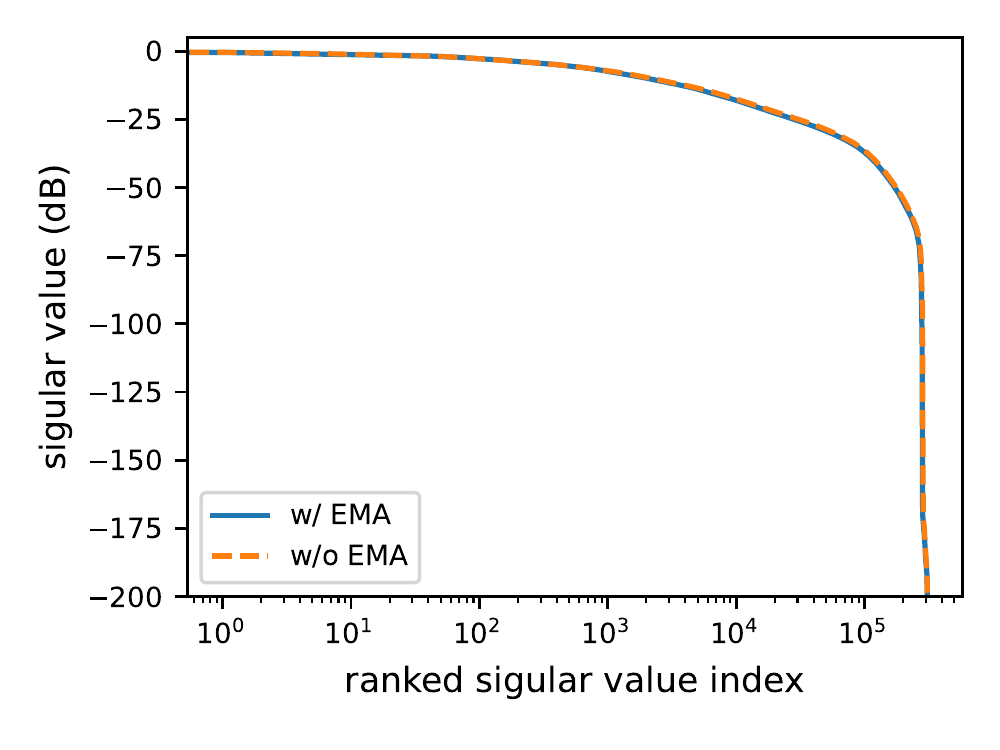}
    \label{fig:subspace_channel_v}
  } \\
  \subfloat[\footnotesize spatial subspace - radio]{
    \includegraphics[width=0.4\textwidth]{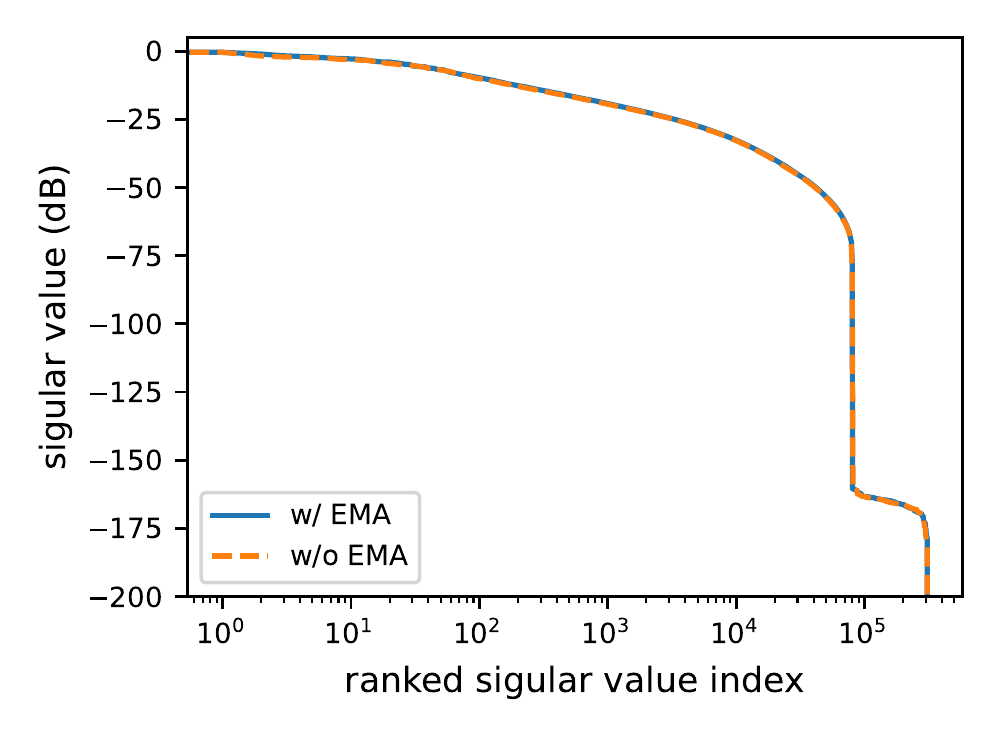}
    \label{fig:subspace_spatial_r}
  }
  \hspace{-0.25cm}
  \subfloat[\footnotesize spatial subspace - visual]{
    \includegraphics[width=0.4\textwidth]{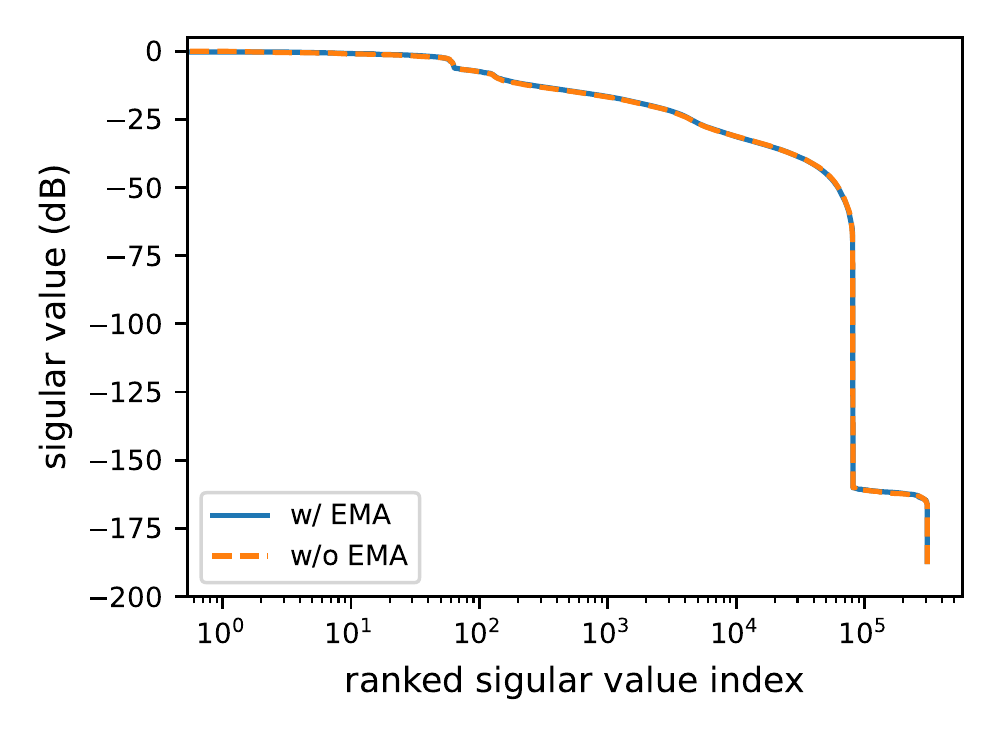}
    \label{fig:subspace_spatial_v}
  }
  \vspace{-0.0cm}
  \caption{Radio-visual subspace analysis w/ and w/o EMA.} 
  \label{fig:radio-visual_subspace_analysis}
\end{minipage}
\end{figure}

In Sec.~\ref{sec:representation_learning}, the spatial encoders $f_{\theta^r}(r), f_{\theta^v}(v) \in \mathbb{R}^{C \times h \times w}$ are introduced.
Following the implementation conditions detailed in Appendix~\ref{sec:implement_details}, $f_{\theta^r}(r), f_{\theta^v}(v)$ are concretely $\in \mathbb{R}^{128 \times 60 \times 80}$. %
In this section, we analyse their dimensionality after 100 epochs of training on the contrastive loss of Eq.~\ref{eq:con_loss}.
To do so, we evaluate these embedding tensors for the validation set.
For each channel $c \in \{1, \dots, 128\}$ and spatial bin $n \in \{1, \dots, 60\}\times\{1, \dots, 80\}$, we compute the centred covariance matrices $\mathrm{Cov}_c \in \mathbb{R}^{128 \times 128}$, $\mathrm{Cov}_{n} \in \mathbb{R}^{4800 \times 4800}$ according to
\begin{align}
  \mathrm{Cov}_x &= \frac{1}{N} \sum_{k=0}^{N-1} (\textbf{z}_k^x - \bar{\textbf{z}}^x)(\textbf{z}_k^x - \bar{\textbf{z}}^x)^T
  \label{eq:covariance}
\end{align}
where $\textbf{z}_k^x$ is the embedding vector of a channel or spatial bin\footnote{i.e., unfolding the original 2-D spatial bins into a vector of $wh=4800$ length} $x \in [c, n]$, $N$ is the number of validation samples, and $\bar{\textbf{z}}^x$ is the respective average.
To measure subspaces dimensionality, we compute the singular value decomposition on the covariance matrix $\mathrm{Cov}_x = U \Sigma V^T, \Sigma=\mathit{diag}(\sigma^k)$, following general practice in SSL theory~\cite{jing2021understanding,tian2021understanding}.
We use these subspace measurements to quantify changes in the learnt contrastive representation as a result of architectural tweaks such as EMA.

We concatenate the singular values of all channels and all spatial bins and sort them in descending order.
Fig.~\ref{fig:radio-visual_subspace_analysis} depicts on a logarithmic scale these aggregated singular values.
We can readily see that EMA has little effect on the dimensionality of the learnt representation across channels and spatial bins, for both radio and vision branches.
We, therefore, opt to exclude it from our experiments for efficiency.

\newpage

\section{Contrastive learning background \& definitions} \label{sec:cl_background} %

\begin{figure}[b]
\begin{minipage}{\textwidth}
  \centering
  \captionsetup{justification=centering}
  \subfloat[\footnotesize CL]{
    \includegraphics[height=2.5cm]{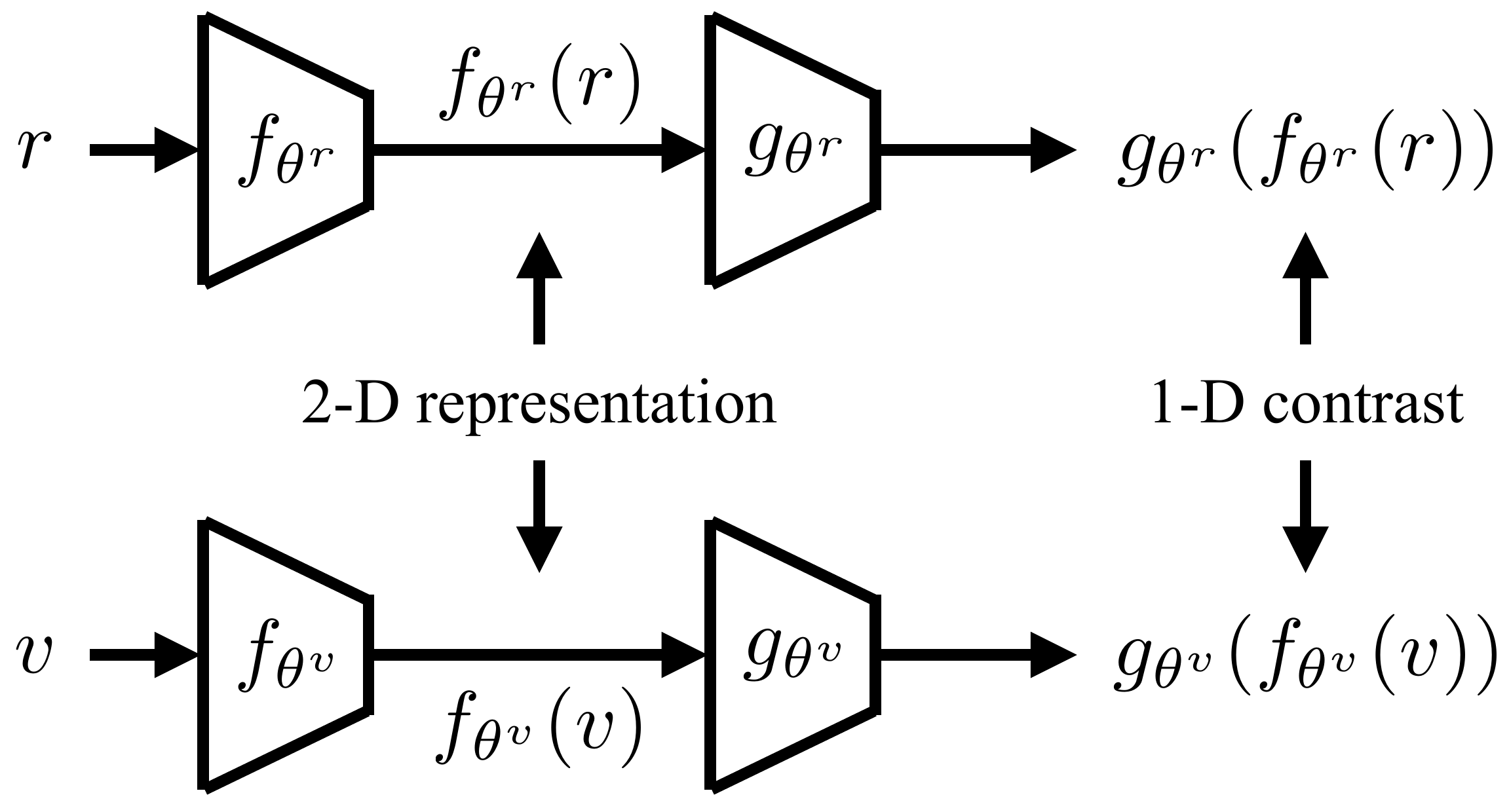} 
    \label{fig:cl_arch}
  }
  \hspace{-0.0cm}
  \subfloat[\footnotesize MCL]{
    \includegraphics[height=2.5cm]{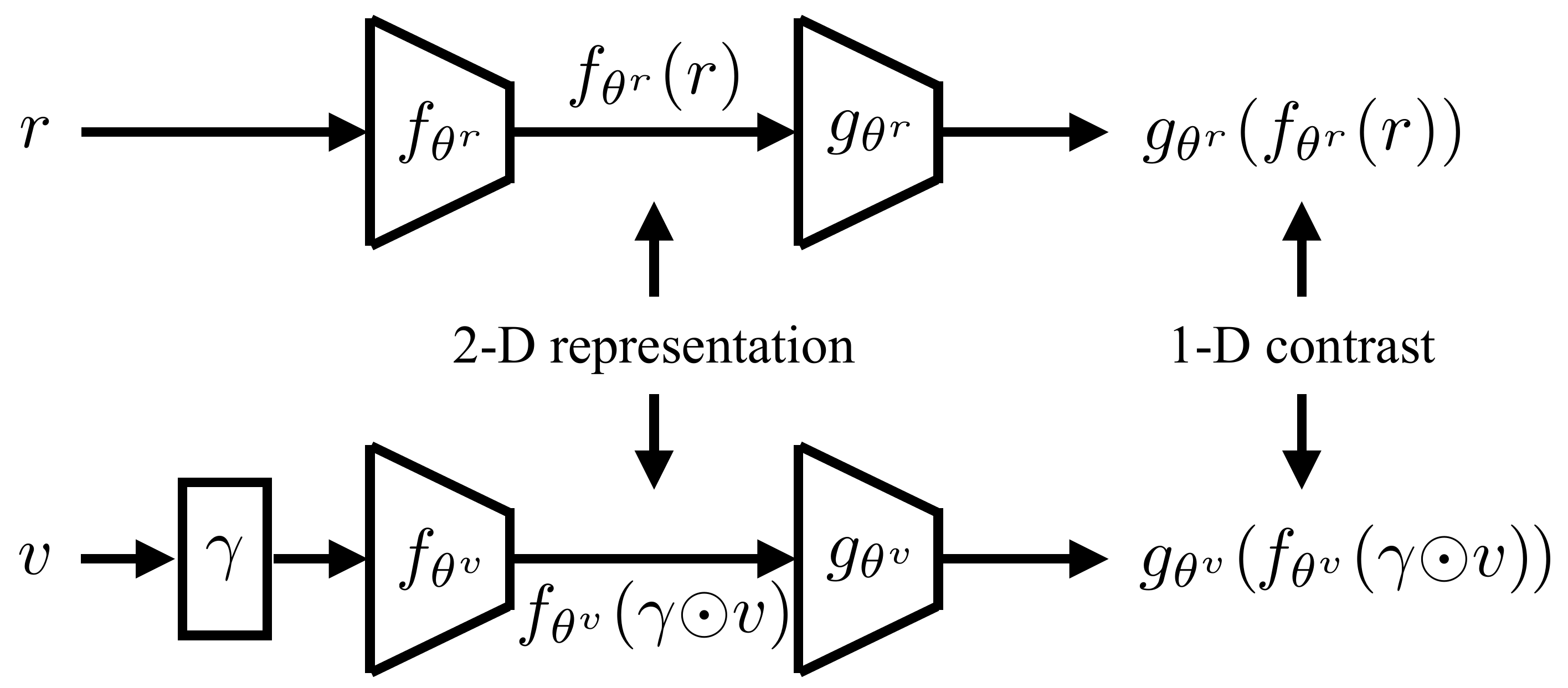}
    \label{fig:mcl_arch}
  } 
  \vspace{-0.20cm}
  \subfloat[\footnotesize SCL]{
    \includegraphics[height=2.5cm]{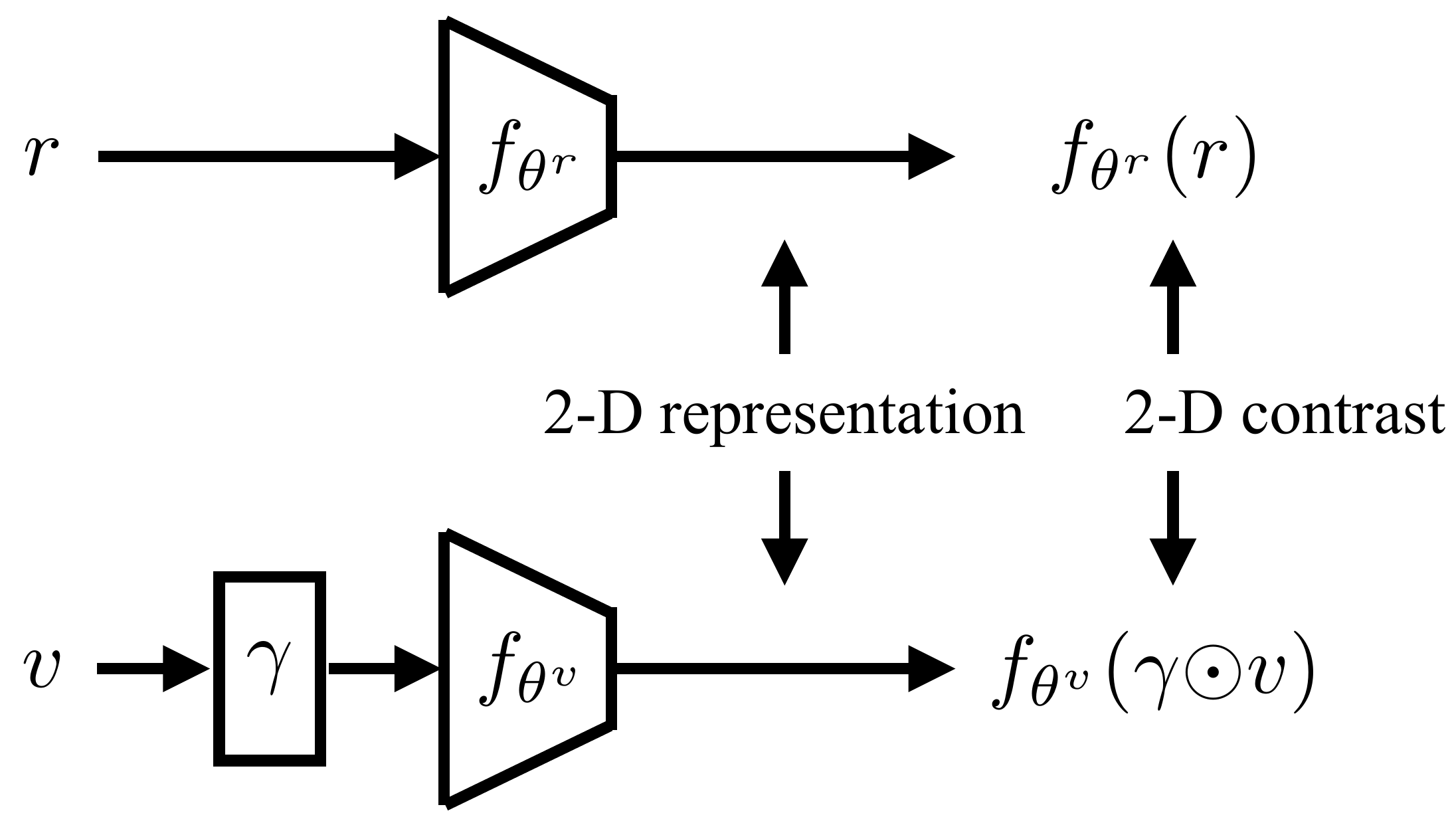}
    \label{fig:scl_arch}
  }
  \caption{\raggedright Three contrastive architectures that use spatial backbones: CL, MCL, and SCL. CL follows the original SimCLR architecture~\cite{chen2020simple} and its accessible queue-based MoCo optimisation~\cite{chen2020improved}, with the addition of a spatial backbone~\cite{chatfield2014return,arandjelovic2018objects}. MCL is broadly similar to CL except for target masking on the vision branch, which promotes added target sensitivity. SCL does not use a projector head and instead rely on spatial contrast~\cite{afouras2020self,afouras2021self,windsor2021self}.}
  \label{fig:contrastive_architectures}
\end{minipage}
\end{figure}

\noindent \textbf{Contrastive learning (CL).} 
Let $(r, v)$ be a radio-visual data pair, where $r \in \mathbb{R}^{1 \times H \times W}$ is a radar heatmap and $v \in \mathbb{R}^{3 \times H \times W}$ is a corresponding RGB image.
Encode, respectively, radio and vision by two neural networks $f_{\theta^r}$ and $f_{\theta^v}$ and their momentum-filtered versions $f_{\bar{\theta}^r}$ and $f_{\bar{\theta}^v}$, assuming some weight parametrisation $\{\theta^r, \theta^v\}$.
Additionally, use projector heads $g_{\theta^r}$ and $g_{\theta^v}$ respectively, such that
\begin{align}
  q^r &= g_{\theta^r}(f_{\theta^r}(r)), \hspace{0.25cm} k^v = g_{\theta^v}(f_{\bar{\theta}^v}(v)), \notag \\ 
  q^v &= g_{\theta^v}(f_{\theta^v}(v)), \hspace{0.25cm} k^r = g_{\theta^r}(f_{\bar{\theta}^r}(r))
  \label{eq:x-modal_encoders}
\end{align}
where vectors $q^r, q^v, k^v, k^r \in \mathbb{R}^{N}$, superscripts $r$ and $v$ denote respectively radio and vision, and following MoCo's query $q$ and key $k$ notation~\cite{he2020momentum}.
With each $r$, use $K+1$ samples of $v$ of which one sample $v^{+}$ is a true match to $r$ and $K$ samples $\{v_{i}^{-}\}_{i=0}^{K-1}$ are false matches---vice versa with each $v$, $K+1$ samples of $r$.
The one-sided cross-modal contrastive losses that test for vision-to-radio and radio-to-vision correspondences are
\begin{align}
  \mathcal{L}_c^{v \rightarrow r}(q^r, k^{{v}+}, \mathbf{k}^{{v}-}) &= - \underset{r,v}{\mathbb{E}} \log \frac{ e^{\operatorname{sim}(q^r, k^{{v}+})} }{ e^{\operatorname{sim}(q^r, k^{{v}+})} +\sum\nolimits_{i} e^{\operatorname{sim}(q^r, k_{i}^{{v}-})} } \notag \\
  \mathcal{L}_c^{r \rightarrow {v}}(q^{v}, k^{r+}, \mathbf{k}^{r-}) &= - \underset{r,v}{\mathbb{E}} \log \frac{ e^{\operatorname{sim}(q^{v}, k^{r+})} }{ e^{\operatorname{sim}(q^{v}, k^{r+})} +\sum_{ i } e^{\operatorname{sim}(q^{v}, k_{i}^{r-})} } \notag
\end{align}
where $\operatorname{sim}(x, y) \coloneqq x^\top y/ \tau$ is a similarity function, $\tau$ is a temperature hyper-parameter, $k^{x+/-} = g_{\theta^x}(f_{\bar{\theta}^x}(x^{+/-}))$ are encodings that denote true and false corresponding signals $x \in [r, v]$ , and vector $\mathbf{k}^{x-} = \{k_{i}^{x-}\}_{i=0}^{K-1}$ holds $K$ false encodings.
Then the bidirectional cross-modal contrastive loss is
\begin{align}
  \mathcal{L}_{\text{CL}} &= ( \mathcal{L}_c^{v \rightarrow r} + \mathcal{L}_c^{r \rightarrow v} )/2
  \label{eq:con_loss}
\end{align}

\noindent \textbf{Spatial contrastive learning (SCL).} \label{sec:scl}
Let $(r, v)$ be a radio-visual data pair, where $r \in \mathbb{R}^{1 \times H \times W}$ is a radar heatmap and $v \in \mathbb{R}^{3 \times H \times W}$ is a corresponding RGB image.
Encode, respectively, radio and vision by two backbone neural networks $f_{\theta^r}$ and $f_{\theta^v}$, assuming some weight parametrisation $\{\theta^r, \theta^v\}$.
Each backbone network encodes per bin one $C$-dimensional feature vector within 2-dimensional spatial bins, i.e., $f_{\theta^r}(r), f_{\theta^v}(v) \in \mathbb{R}^{C \times h \times w}$. 
The spatial binning resolution $h \times w$ is generally coarser than the original image resolution $H \times W$.
Denote by $f^{r}_{n}(r), f^{v}_{n}(v) \in \mathbb{R}^{C}$ radio and vision spatial encodings at bin $n \in \Omega = \{1, \dots, h\} \times \{1, \dots, w\}$.
Construct a target mask $\gamma \coloneqq [ \gamma_{ij} ] \in [0, 1]^{H \times W}$ such that $f^{\tilde{v}}_{m}(\gamma \odot v) \in \mathbb{R}^{C}$ is defined for $m \in \tilde{\Omega} = \{1, \dots, \tilde{h}\} \times \{1, \dots, \tilde{w}\}$ to retain encodings for the target of interest only in the RGB image (e.g., as delineated by a bounding box), where $\odot$ is the element-wise product and $\tilde{\Omega} \subset \Omega$ is a subset of spatial locations.
In practice, the target mask can either be (1) estimated using off-the-shelf vision object detectors such as Yolo~\cite{redmon2016you,glenn_jocher_2021_5563715}, or (2) obtained directly as groundtruth during data synthesis.

Noting attention maximisation defined earlier in main paper in Eqs.~\ref{eq:2d_attention}~\&~\ref{eq:max_2d_attention}, spatial cross-modal contrastive losses can then be implemented during training from a batch $\mathcal{B}$ for all radio-visual pairs $(r, v) \in \mathcal{B}$ according to
\begin{align}
  \mathcal{L}_a^{v \rightarrow r}(\mathcal{B}) &= - \underset{\mathcal{B}}{\mathbb{E}} \; \log \frac{ \exp{\big( S(r, v)/\tau \big)} }{ \sum\limits_{ i \in \mathcal{B} } \exp{\big(  S(r, v_i)/\tau \big)} }, \notag \\  
  \mathcal{L}_a^{r \rightarrow v}(\mathcal{B}) &= - \underset{\mathcal{B}}{\mathbb{E}} \; \log \frac{ \exp{\big( S(r, v)/\tau \big)} }{ \sum\limits_{ i \in \mathcal{B} } \exp{\big(  S(r_i, v)/\tau \big)} } 
  \label{eq:x-modal_spatial_contrastive_loss}
\end{align}
where the one-sided loss $\mathcal{L}_a^{v \rightarrow r}$ tests for vision-to-radio correspondence, similarly $\mathcal{L}_a^{r \rightarrow v}$ tests for radio-to-vision, and $\tau$ is a temperature hyper-parameter.
The bidirectional contrastive loss that incentivises cross-modal spatial attention becomes
\begin{align}
  \mathcal{L}_{\text{SCL}} &= ( \mathcal{L}_a^{v \rightarrow r} + \mathcal{L}_a^{r \rightarrow v} )/2
\end{align}

For clarity, Fig.~\ref{fig:contrastive_architectures} illustrates the three contrastive learning flavours used in this work.

\newpage

\section{Implementation details} \label{sec:implement_details}

\vspace{-0.0cm}
The spatial backbone of the radio and vision encoders uses an architecture similar to VGG-M~\cite{chatfield2014return,arandjelovic2018objects}, swapping max pooling for average pooling as recommended in~\cite{afouras2021self}.
For standard contrastive ablation in Sec.~\ref{sec:ablations_analysis} (Contrastive Learning (CL) \& Masked Contrastive Learning (MCL)), we base our cross-modal contrastive learning on MoCo v2 and its public implementation~\cite{he2020momentum}. 
We extend MoCo's implementation with two queues for radio and vision similar to the audio-visual active sampling work in~\cite{ma2020active}.
We have found that filtering the encoders with exponential moving average (EMA) when implementing radio-visual contrast has no tangible advantage, as detailed in Appendix~\ref{sec:appendix_subspace}.

For mask generation in vision, we rely on groundtruth bounding boxes from Blender.  
We also characterise downstream performance using bounding boxes estimated from off-the-shelf Yolov5 model~\cite{glenn_jocher_2021_5563715}.
Tab.~\ref{tab:yolov5_perf_maxray} reports Yolov5's IoU-$0.5$ performance metric as measured on~\dataset.

We train on 640$\times$480 resolution for both RGB images and radio heatmaps.
Both radio and vision branches output 128$\times$80$\times$60 spatial features whose dimensionality is reduced using 2-layer MLP projectors to 64-D vectors in the case of CL \& MCL.
For CL \& MCL, we use a MoCo v2 queue whose size equals to the batch size.
For CL \& MCL, the temperature hyper-parameter is 0.07, whereas for SCL it is 0.1.
When implementing spatial attention, we pad bounding boxes by a margin of 5 pixels, and pad a target spatial response by a margin of 1 feature.
For backbone training, we use the Adam optimiser~\cite{kingma2014adam} with a learning rate of $10^{-5}$ and no schedule.
For all model variants, we train for 200 epochs. 
We use a batch size of 32 and train in a distributed fashion on 8 GPUs.
We trained experiments on two machines with GeForce RTX 2080 Ti GPUs and RTX A5000 GPUs, throughout for backbone training, supervised training, and NNI search space.
Backbone training takes around 16--24 hours per experiment depending on model variant and configuration.
Both the localiser network trained on self-coordinates and supervised baseline use identical architecture and training as detailed in Tab.~\ref{tab:NNIparameter}.
\dataset~and CRUW use different convolutional network settings due to differences in range and angular resolutions (cf., Tab.~\ref{tab:maxray_vs_cruw}).
The NNI search space took around 5 days.
For~ray tracing~\dataset, the ray casting settings of Blender greatly influence performance. 
We set the maximum number of interactions to 5 and the maximum length travelled to 500m. 
We parallelise frame creation on 3060 Ti GPU, which gives 200sec creation time per frame. 
This results in a total of 11.6 days of ray tracing time for the parking lot scenario of dataset.

\newpage

\vspace{-0.25cm}
\section{OFDM radar primer} \label{sec:appendix_ofdm_radar}

Sec.~\ref{sec:dataset_modelling_details} detailed the modelling and synthesis flow~\dataset~incorporates for vision and radio data.
6G network design is an active area of research whose details are in a state of flux.
We, therefore, elaborate here on our radio data synthesis flow in order to enhance the clarity of~\dataset's radio modelling and assumptions.

\begin{figure}[b]
\begin{minipage}{\textwidth}
  \centering
  \includegraphics[width=0.75\textwidth]{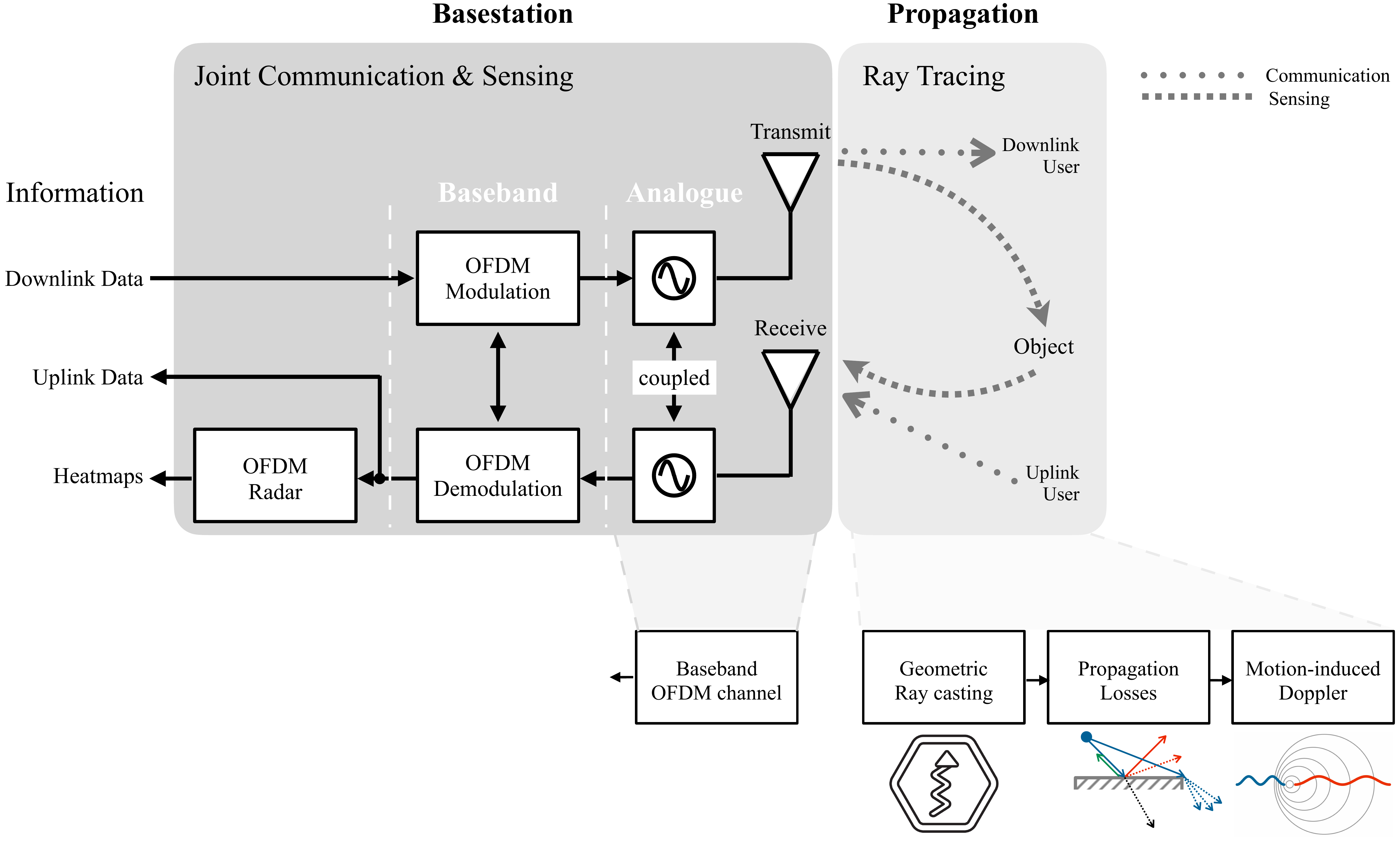}
  \caption{6G network model with sensing support used in~\dataset. The model consists of two subsystems: (a) basestation and (b) propagation. The basestation implements OFDM radar signalling within a phase coherent signal processing architecture. We simulate the OFDM channel in baseband. Propagation simulations are performed via geometric ray tracing. We extensively model propagation losses (e.g., diffraction, backscatter, reflection, scatter, penetration, etc.) as well as Doppler effects.}
  \label{fig:6g_sim}
\end{minipage}
\end{figure}

Fig.~\ref{fig:6g_sim} depicts a simplified block diagram of our 6G cellular system with sensing support.
This 6G model consists of two simulation flows: (a) propagation via ray tracing, and (b) OFDM-based basestation signal processing.

\noindent \textbf{(a) Propagation.} 
The basestation transmits OFDM signals. 
These OFDM signals interact with the synthetic environment of Blender through a set of complex propagation phenomena.
As such, backscatter signals captured at the basestation receiver chain enable radar detection.
For synthesising these backscatter signals,~\dataset~uses high-fidelity radio ray tracing.
Specifically,~\dataset~(i) implements geometric radio ray casting within Blender, (ii) calculates the propagation losses of these rays upon interacting with the synthetic environment model, and (iii) induces approperiate Doppler effects that correspond to moving objects (see Fig.~\ref{fig:6g_sim}).
The propagation model concludes by presenting ``environmentally-modulated'' OFDM signals back to the basestation model.

\noindent \textbf{(b) Basestation.}
In 6G networks, sensing is to be supported at the \emph{physical layer}, unlike earlier attempts for opportunistically using standard wireless channel estimates for sensing~\cite{alloulah2019future}.
For this to happen, the basestation transmits OFDM signals and then receives them back ``modulated'' by environmental effects.
Specifically, the echoes backscattered from objects in the environments are received back at the basestation \emph{coherently} w.r.t. the local oscillator of the receive chain.
This coherent transceiver is illustrated in Fig.~\ref{fig:6g_sim} as a \emph{coupling} between the transmit and receive analogue chains.
The modified transceiver remains compatible with standard downlink and uplink communications.

Radar processing in~\dataset~is then implemented on top of OFDM communication signals.
OFDM is the workhorse of modern communication systems.
Using OFDM radar makes sensing much more amenable to integration in communication systems.
Specifically, OFDM radar processing begins after we obtain wireless channel estimates from the OFDM demodulator as shown in Fig.~\ref{fig:6g_sim}.
OFDM radar finally outputs the sensing primitives (i.e., the heatmaps) that our radio-visual SSL uses.

Note that joint communication and sensing in 6G as illustrated in Fig.~\ref{fig:6g_sim} is non-trivial. 
Concretely, 6G requires (a) new hardware at cellular basestations, as well as (b) new resource allocation protocol involving space, time, frequency, and power optimisations of the network~\cite{wild2021joint}.   
For completeness, the following describes briefly the signal processing principles of OFDM radar~\cite{Braun11_OfdmRadar} as implemented in~\dataset.

\noindent \textbf{OFDM radar signal processing.}
For $N_\text{symb}$ known transmitted symbols $\textbf{X}$, the channel can be estimated from the received data $\textbf{Y}$ according to
\begin{align}
  \textbf{H}^{k,n} = \frac{\textbf{Y}^{k,n}}{\textbf{X}^{k,n}}
\end{align}
where $\textbf{X}, \textbf{Y} \in \mathbb{C}^{N_\text{sub} \times N_\text{symb}}$, $N_\text{sub}$ is the number of subcarriers, $k$ and $n$ are respectively subcarrier and symbol indices, and division is element-wise for efficient single tap equalisation. 
The signal traverses a finite number of paths $L$ to the receiver. 
As such we can write the channel according to
\vspace{-0.20cm}
\begin{align}
  \textbf{H}^{k,n} = \sum_{\ell=0}^{L} \rho_{\text{loss}} \; \underbrace{e^{j2\pi nT_0f_{\ell}}}_{\text{Doppler}} + \; \underbrace{e^{j2\pi k d_{\ell}/c_0 \Delta f}}_{\text{distance}} + \; \eta^{k,n}
  \label{eq:ofdm_radar}
\end{align}
where $f_{\ell}$ is the per-path Doppler-induced phase shift that modulates OFDM symbols, and $T_0$ is the symbol duration. 
The distance travelled induces another phase shift that affects OFDM subcarriers, with $\Delta f$ being the subcarrier spacing, and $\eta^{k,n} \sim \mathcal{N}(0, \sigma^{k,n})$ is zero-mean Gaussian noise. 
Eq.~\eqref{eq:ofdm_radar} tells us that the phase information per path (i.e., bounced off some object) can be used to determine the relative speed and range of objects encountered during propagation. 
The angle of an object can also be estimated by phase processing multiple $\textbf{H}^{k,n}$ across antennae, i.e., spatial processing. 
Orthogonality in OFDM allows for efficient periodogram estimation of the channel as~\cite{Braun11_OfdmRadar}
\vspace{-0.35cm}
\begin{align}
  \textbf{P}^{k,n} = \left| \sum_{m=0}^{N_\text{smyb}-1}\left(\sum_{p=0}^{N_\text{sub}-1}\textbf{H}^{p,m}e^{-j2\pi\frac{pn}{N_\text{symb}}}\right)e^{j2\pi\frac{mk}{N_\text{sub}}}\right|^2
\end{align}
using the fast Fourier transform (FFT) over symbols, and the inverse FFT over subcarriers. 
This gives rise to peaks at the corresponding distance and speed of respective objects.

The above treatment shows that OFDM signalling for communication can be reused for implementing radar techniques for sensing.
Integrating such sensing functionality alongside communications, with acceptable tradeoffs, is an active area of research for 6G networks.

\newpage

\vspace{-0.25cm}
\section{Self-labels analysis} \label{sec:appendix_selfLabel}

\begin{figure}[b]
\begin{minipage}{\textwidth}
  \centering
  \captionsetup{justification=centering}
  \subfloat[\footnotesize SCL range - Train]{
    \includegraphics[trim=0 0 0 0.5cm, clip, width=0.235\textwidth]{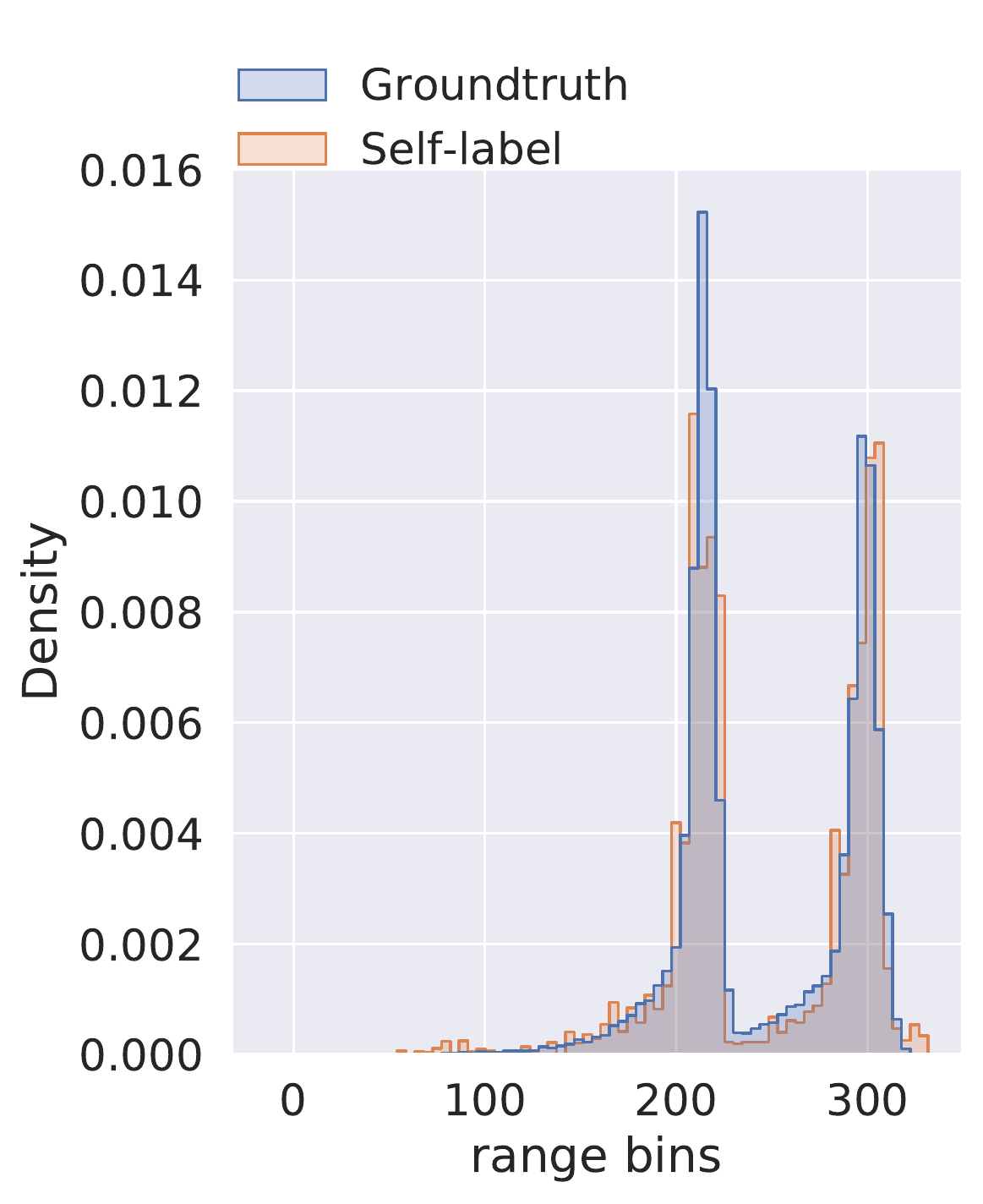}
    \label{fig:range_train_dists_scl}
  }
  \hspace{-0.25cm}
  \subfloat[\footnotesize MCL range - Train]{
    \includegraphics[trim=0 0 0 0.5cm, clip, width=0.235\textwidth]{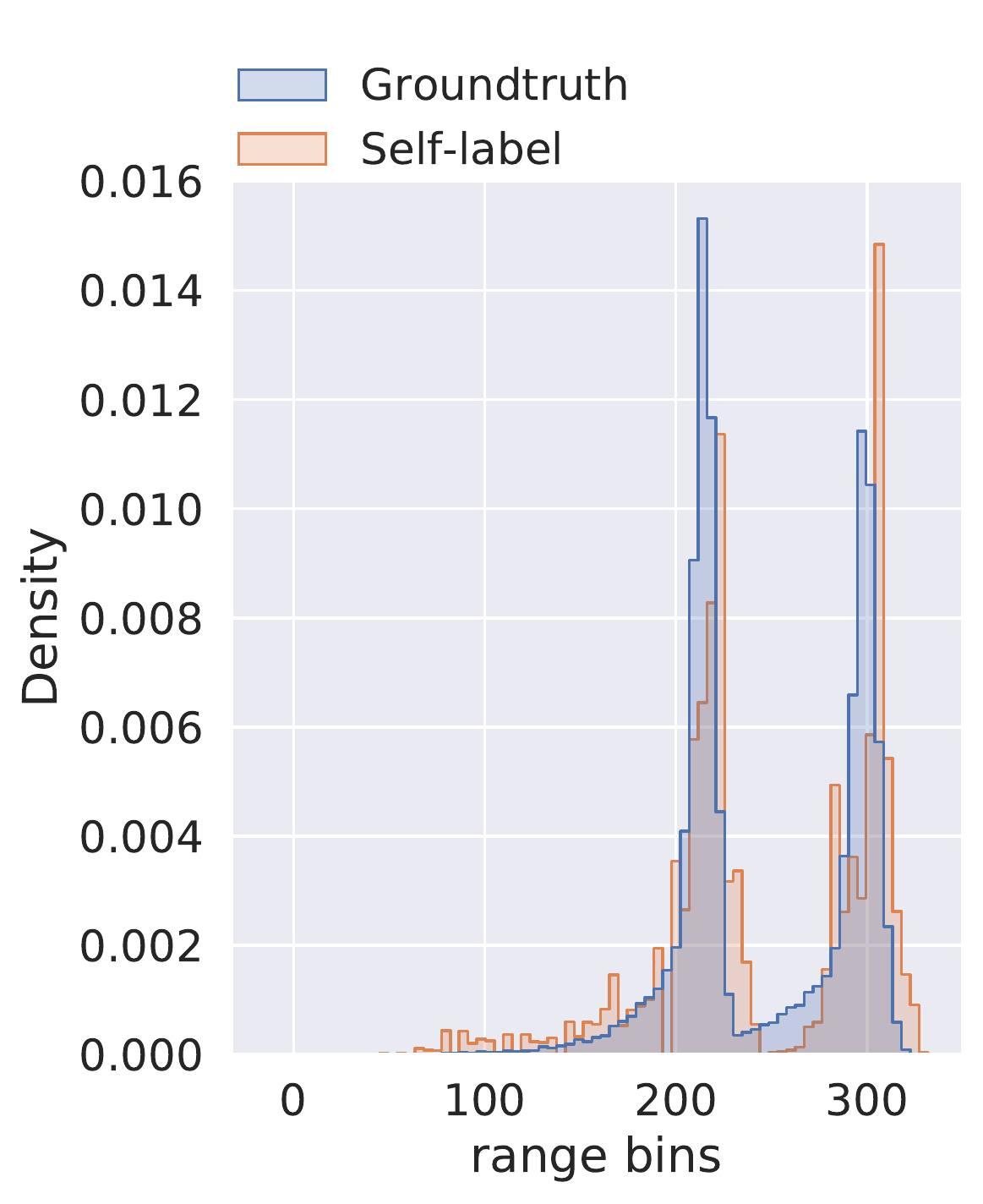}
    \label{fig:range_train_dists_mcl}
  } 
  \hspace{-0.25cm}
  \subfloat[\footnotesize SCL range - Valid]{
    \includegraphics[trim=0 0 0 0.5cm, clip, width=0.235\textwidth]{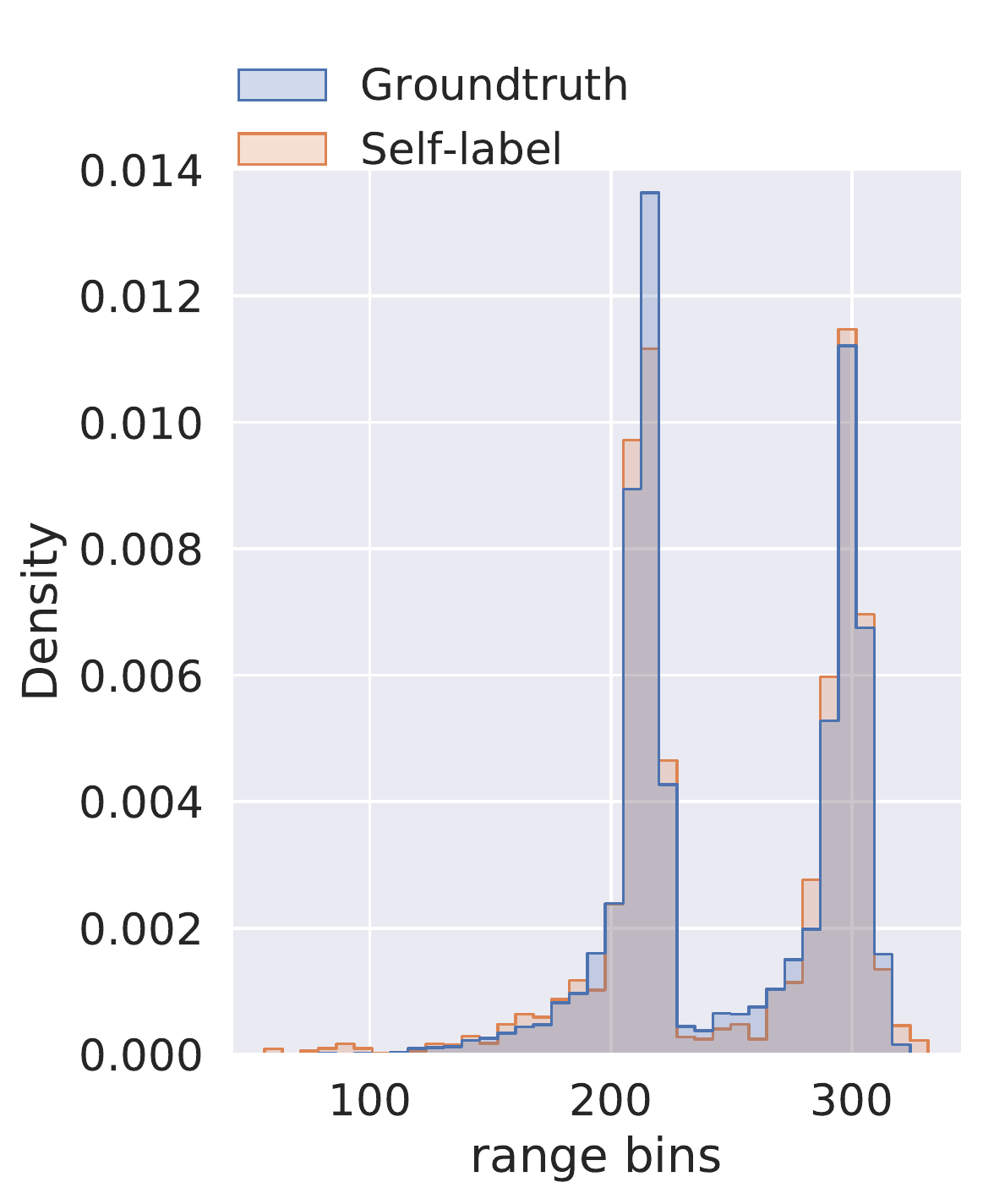}
    \label{fig:range_valid_dists_scl}
  }
  \hspace{-0.25cm}
  \subfloat[\footnotesize MCL range - Valid]{
    \includegraphics[trim=0 0 0 0.5cm, clip, width=0.235\textwidth]{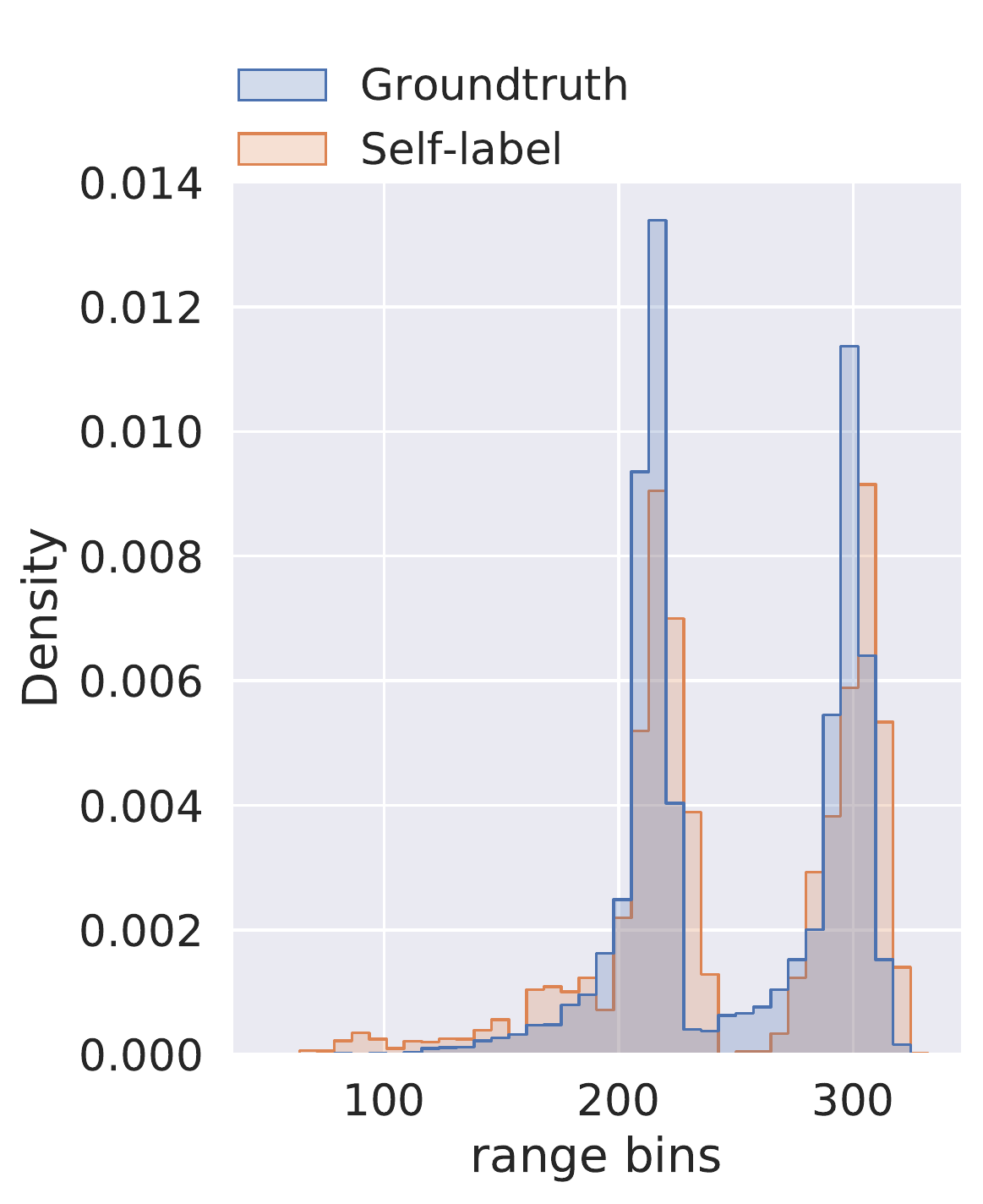}
    \label{fig:range_valid_dists_mcl}
  } \\
  \subfloat[\footnotesize SCL angle - Train]{
    \includegraphics[trim=0 0 0 0.5cm, clip, width=0.235\textwidth]{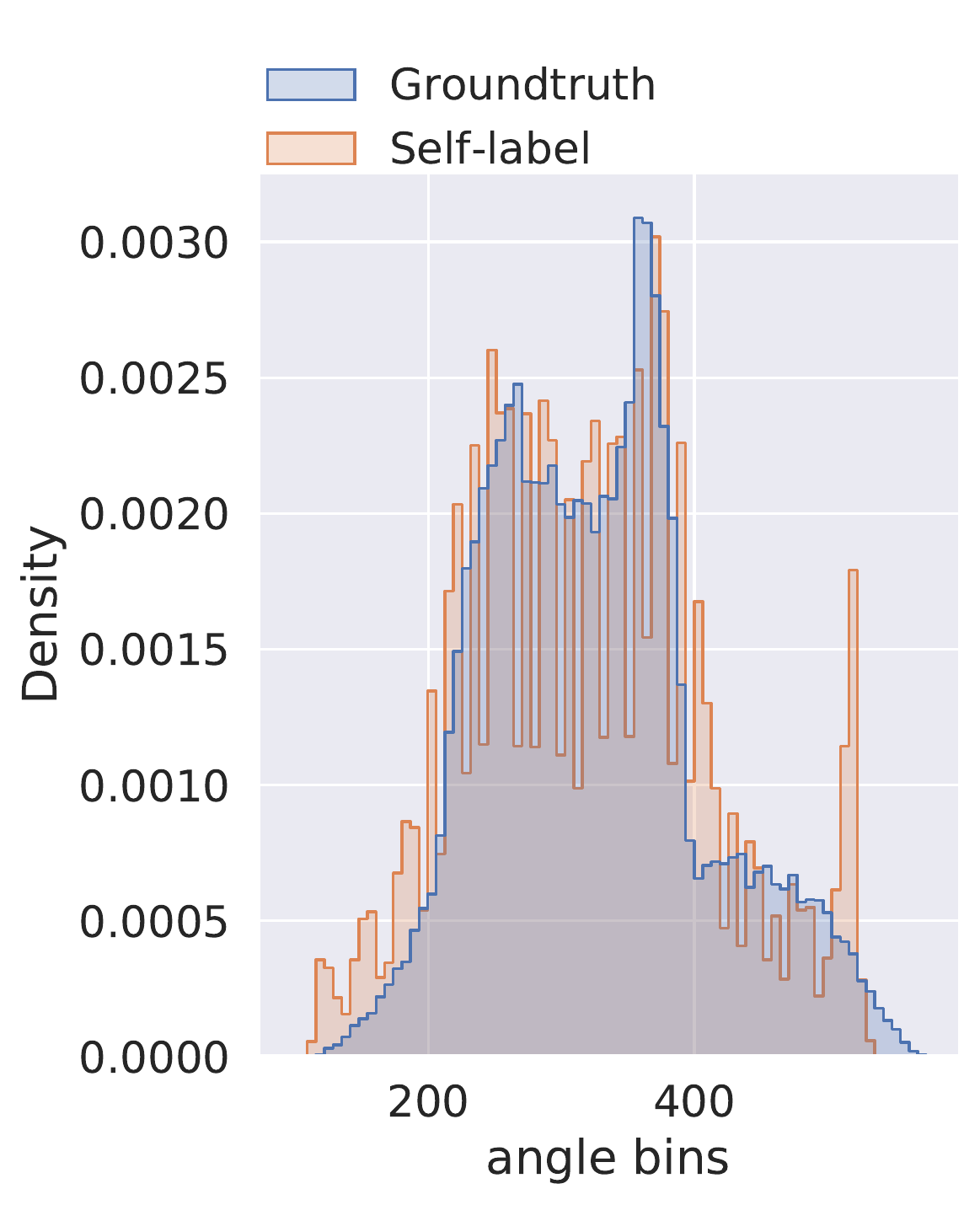}
    \label{fig:azimuth_train_dists_scl}
  }
  \hspace{-0.25cm}
  \subfloat[\footnotesize MCL angle - Train]{
    \includegraphics[trim=0 0 0 0.5cm, clip, width=0.235\textwidth]{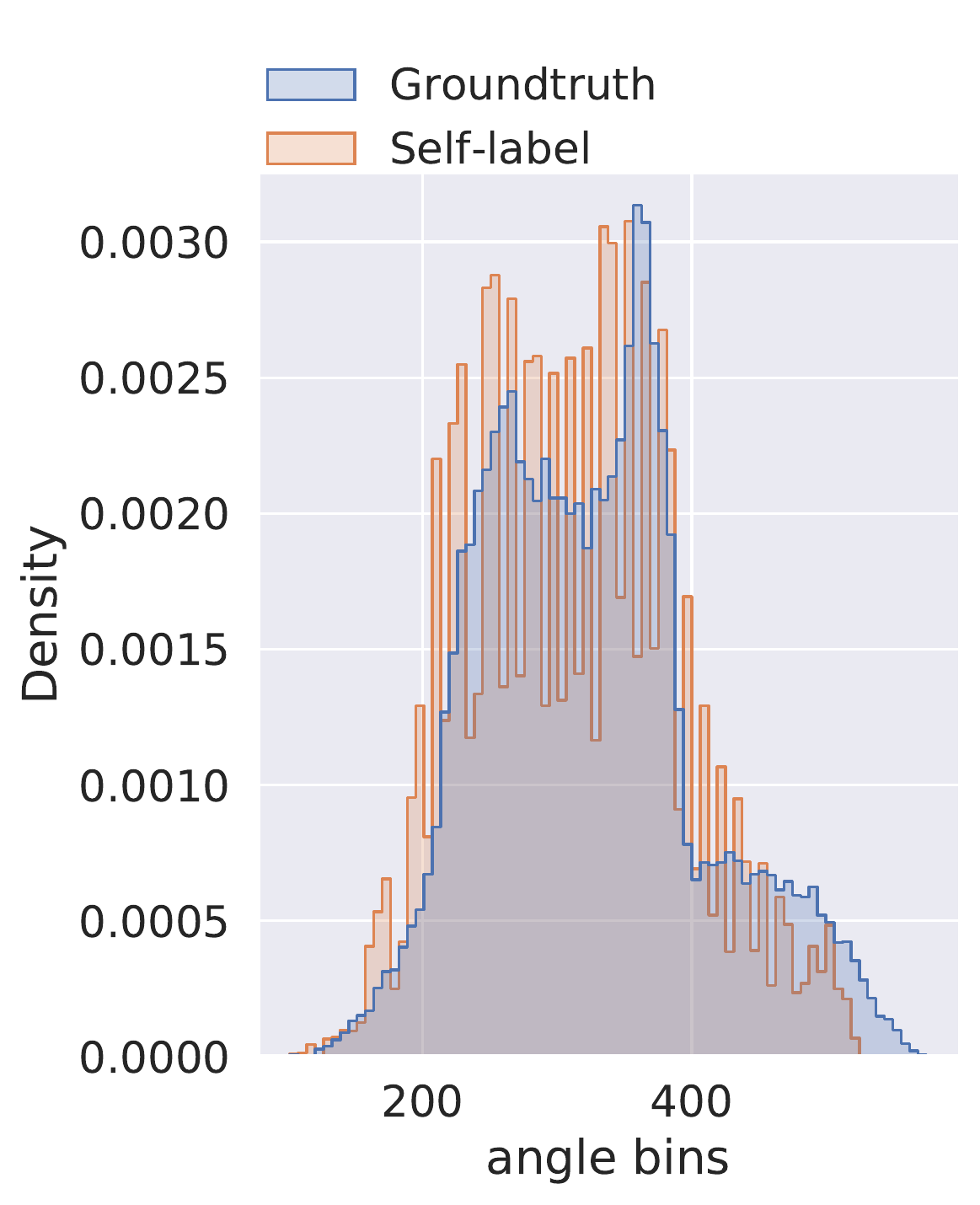}
    \label{fig:azimuth_train_dists_mcl}
  } 
  \hspace{-0.25cm}
  \subfloat[\footnotesize SCL angle - Valid]{
    \includegraphics[trim=0 0 0 0.5cm, clip, width=0.235\textwidth]{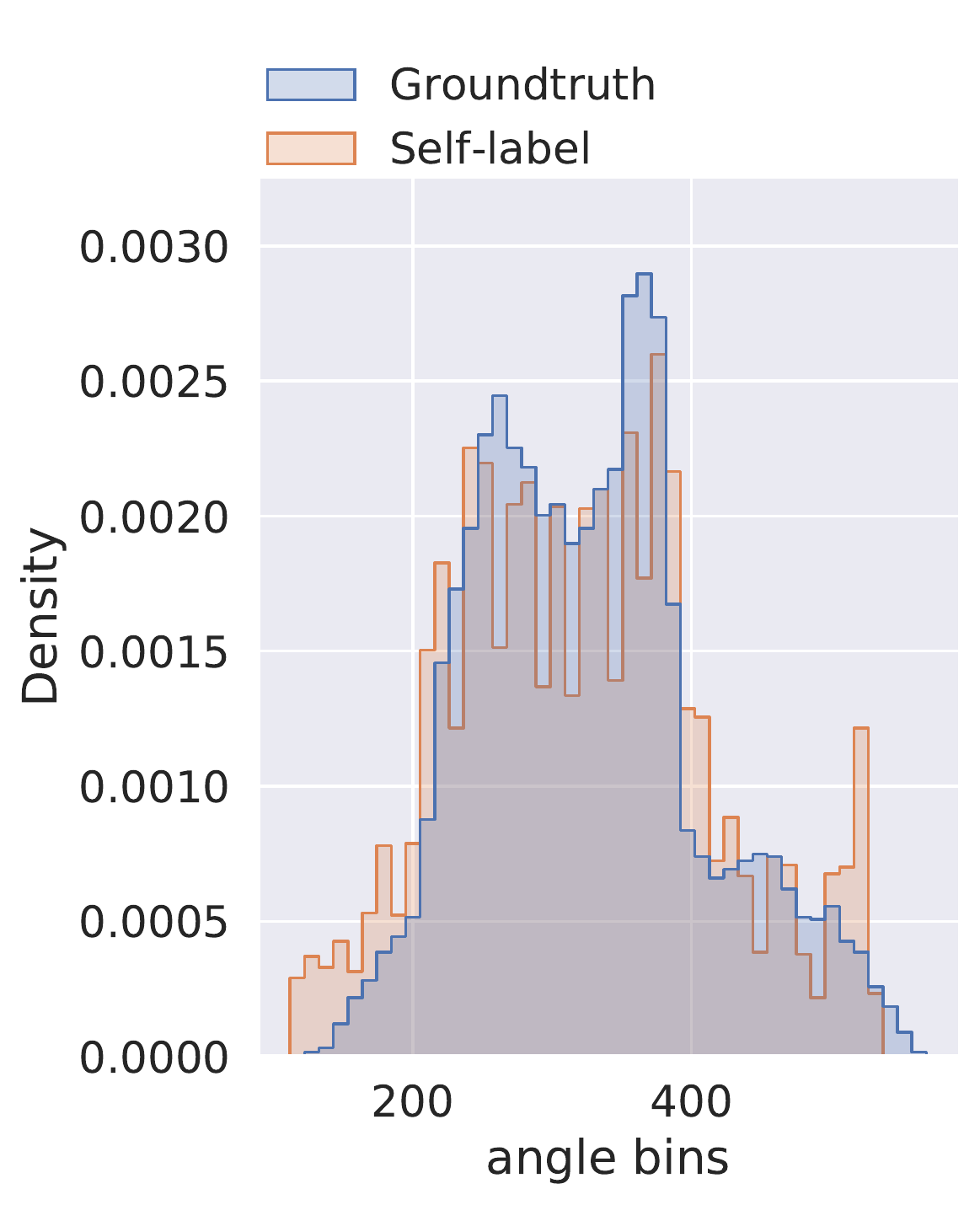}
    \label{fig:azimuth_valid_dists_scl}
  }
  \hspace{-0.25cm}
  \subfloat[\footnotesize MCL angle - Valid]{
    \includegraphics[trim=0 0 0 0.5cm, clip, width=0.235\textwidth]{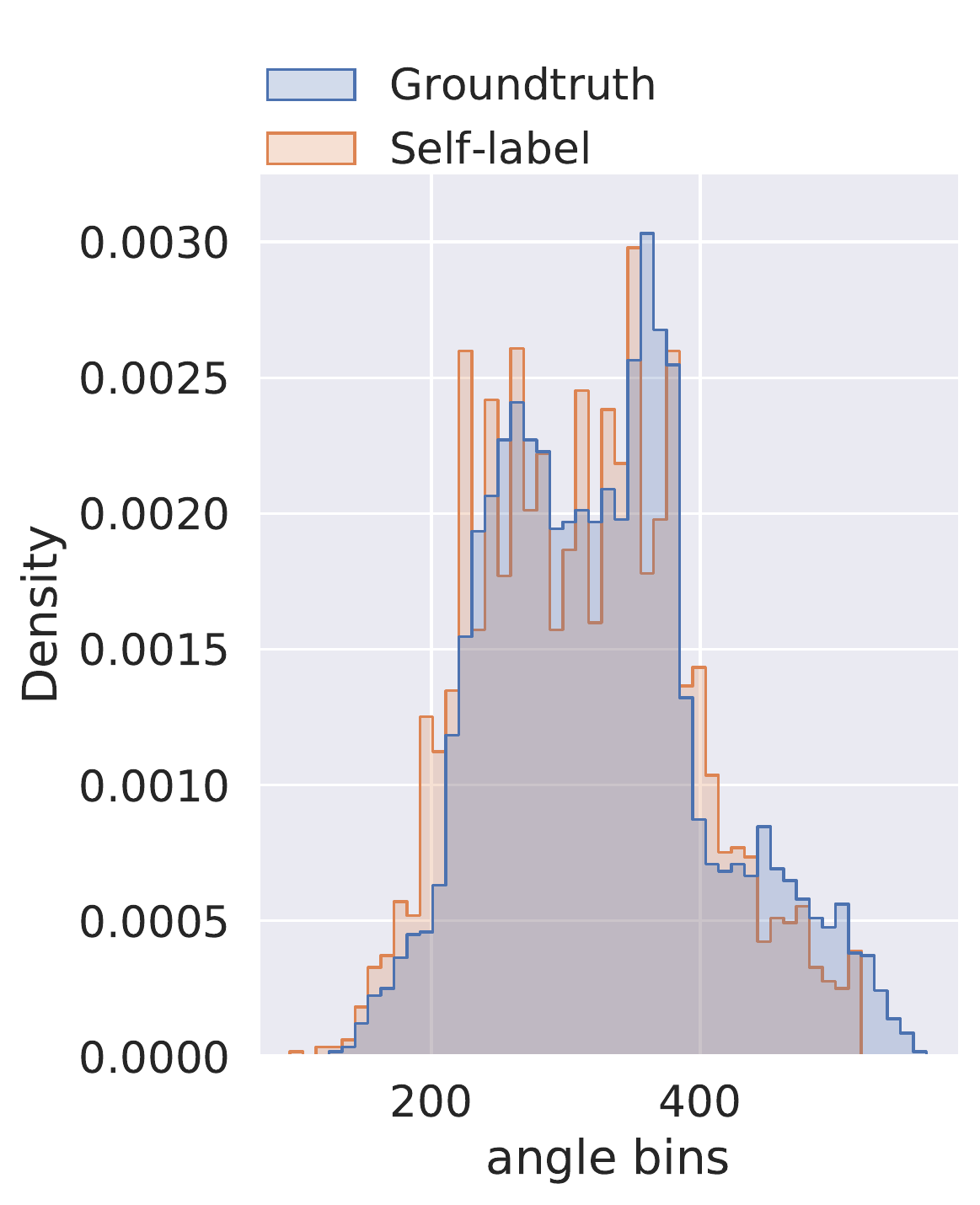}
    \label{fig:azimuth_valid_dists_mcl}
  }
  \vspace{-0.085cm}
  \caption{\raggedright Groundtruth label distributions for target range and angle bins, along with their respective self-label distributions overlaid. SCL and MCL behave differently in their ability to derive self-labels. Such distributions are depicted for both training and validation sets.}
  \label{fig:selfLabels_vs_groundtruth_distributions}
\end{minipage}
\end{figure}

Further to discussions in Sec.~\ref{sec:ablations_analysis}, Fig.~\ref{fig:selfLabels_vs_groundtruth_distributions} details the empirical histograms that characterise SCL's and MCL's self-label deviation from groundtruth labels.

\newpage

\section{Additional results} \label{sec:additional_results}

\vspace{-0.0cm}
\begin{table}[h]
\begin{minipage}[b]{.225\textwidth}
    \centering
    \caption{Yolov5 performance on~\dataset.}
    \medskip
    \vspace{-0.40cm}
  \scriptsize
  \setlength{\tabcolsep}{3pt}
  \begin{tabular}{lcc}
    \toprule
     
                         & mAP$_{50}$  & IoU-$0.5$  \\
    \midrule
    \dataset             & 100         & 0.9374     \\ 
    \bottomrule
    \label{tab:yolov5_perf_maxray}
  \end{tabular}
\end{minipage}\hfill
\begin{minipage}[b]{.225\textwidth}
    
\end{minipage}
\end{table}

\newpage

\section{MaxRay illustrations} \label{sec:appendix_maxray_illustra}

\begin{figure}[b]
\begin{minipage}{\textwidth}
  \centering
  \captionsetup{justification=centering}
  \subfloat[\footnotesize Camera]{
    \includegraphics[height=2.75cm]{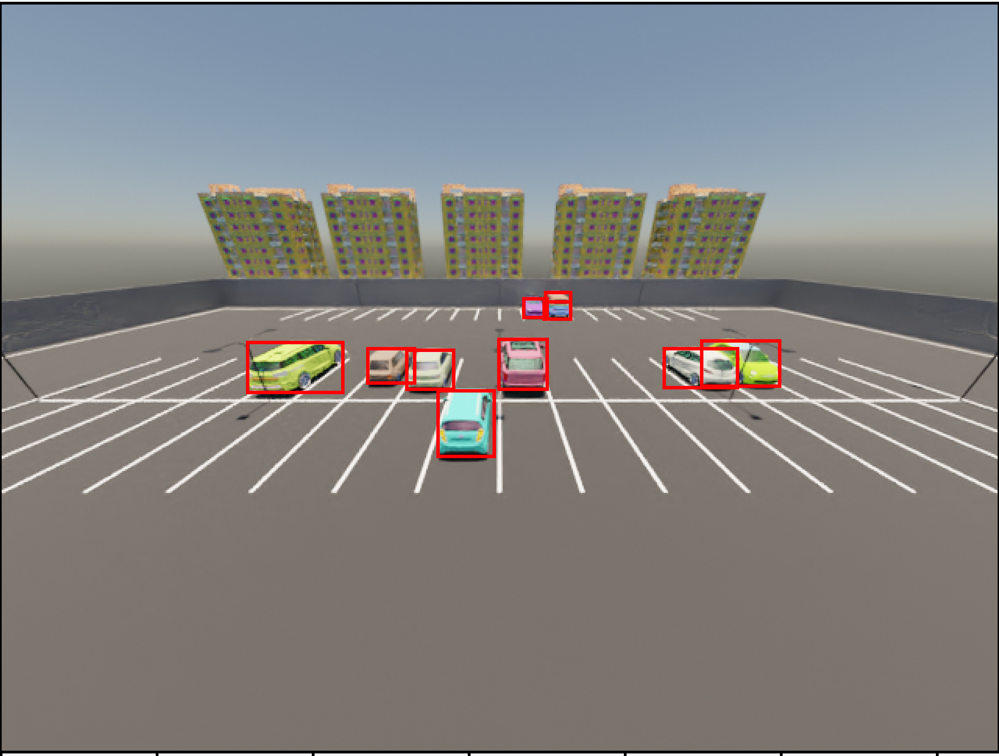} 
  }
  \hspace{-0.25cm}
  \subfloat[\footnotesize Lidar]{
    \includegraphics[height=2.75cm]{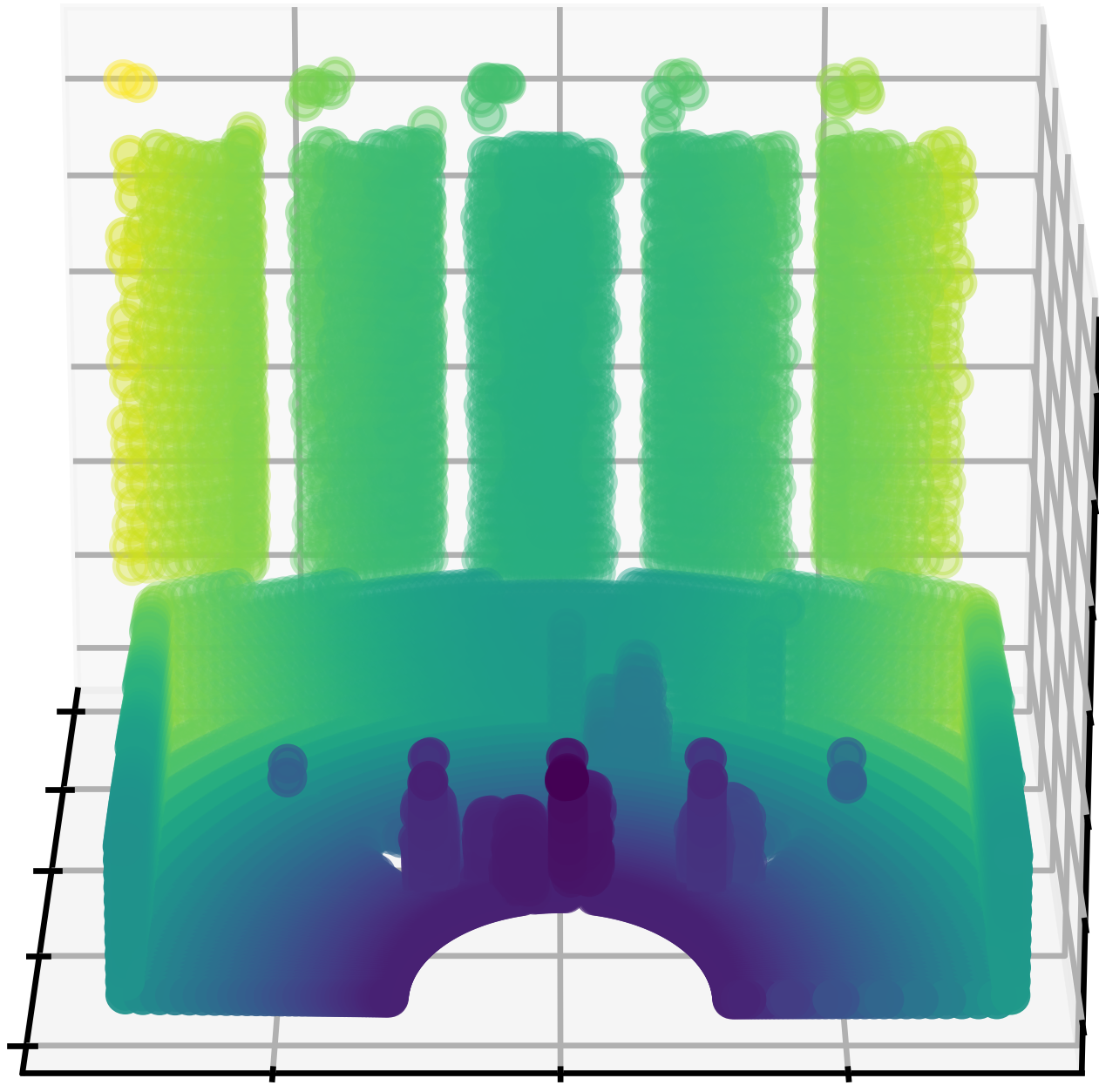}
  }
  \hspace{-0.25cm}
  \subfloat[\footnotesize Depth]{
    \includegraphics[height=2.75cm]{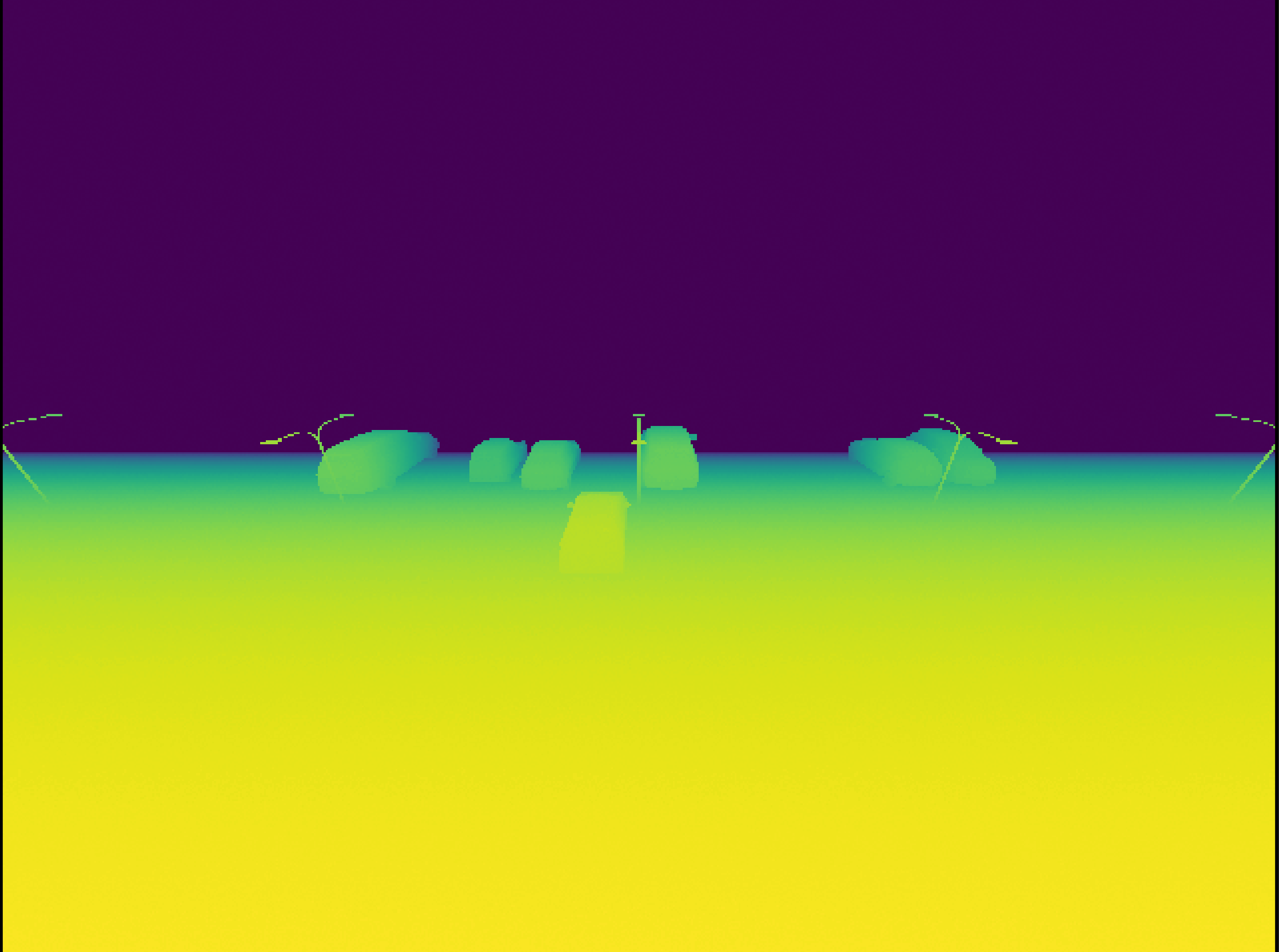} 
  }
  \hspace{-0.25cm}
  \subfloat[\footnotesize Radar]{
    \includegraphics[height=2.75cm]{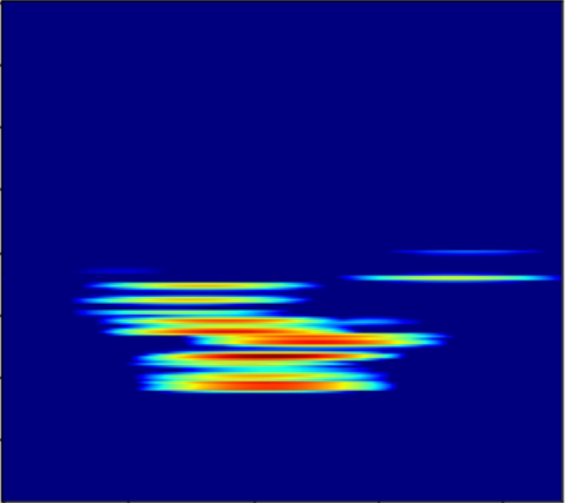}
  } \\
  \subfloat[\footnotesize Car dimension statistics]{
    \includegraphics[height=3.75cm]{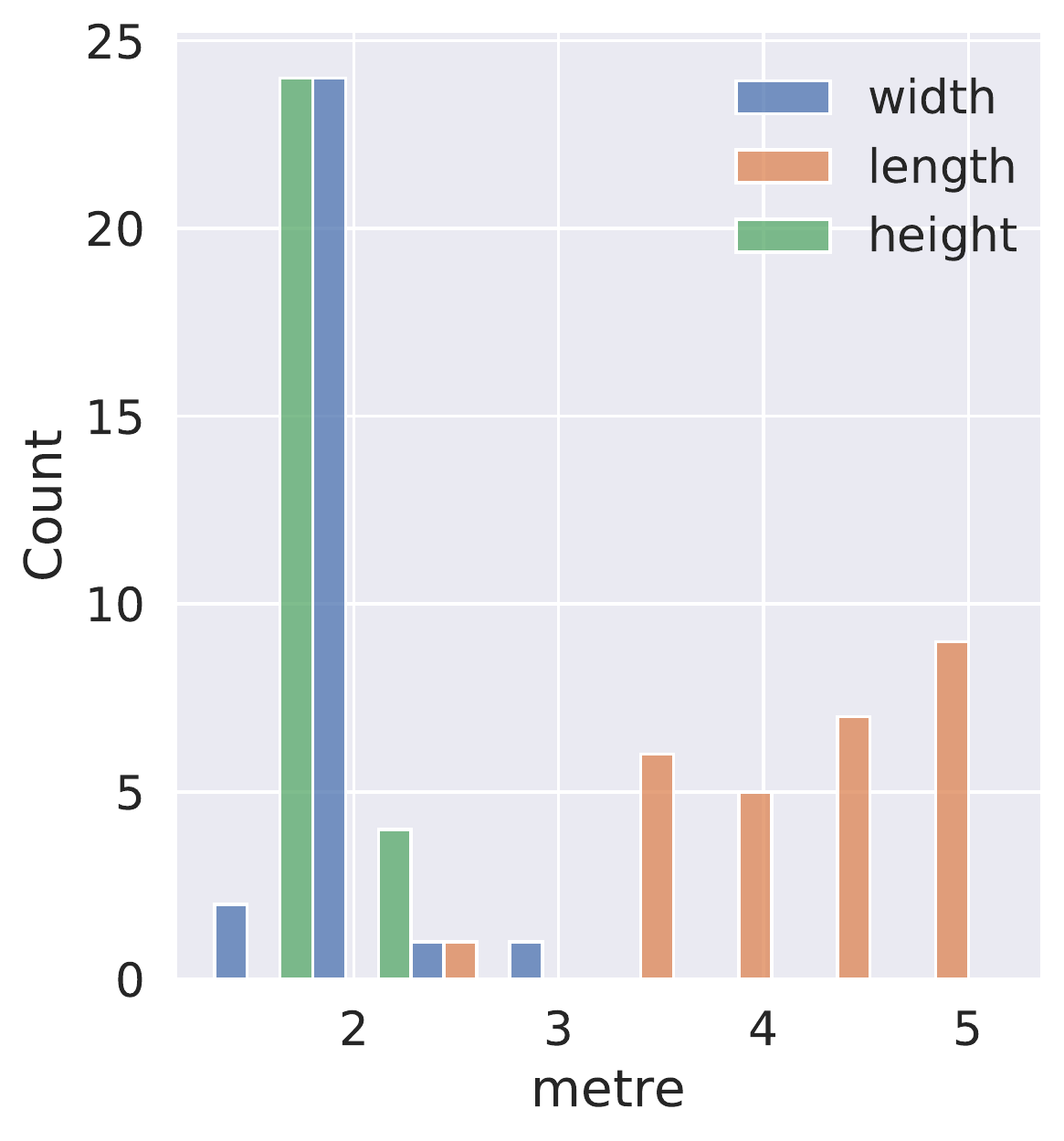}
    \label{fig:parkingLot_car_stats}
  }
  \caption{\raggedright Example of different modalities supported in~\dataset. From left to right: Camera image with bounding boxes, Lidar point cloud with object type, Depth image with range, Radar heatmap with groundtruth coordinates. Distribution of car dimensions throughout dataset is also illustrated.}
  \label{fig:maxray_examples}
\end{minipage}
\begin{minipage}{\textwidth}
\vspace{0.75cm}
  \centering
  \captionsetup{justification=centering}
  \subfloat[\footnotesize Day]{
    \includegraphics[height=3.75cm]{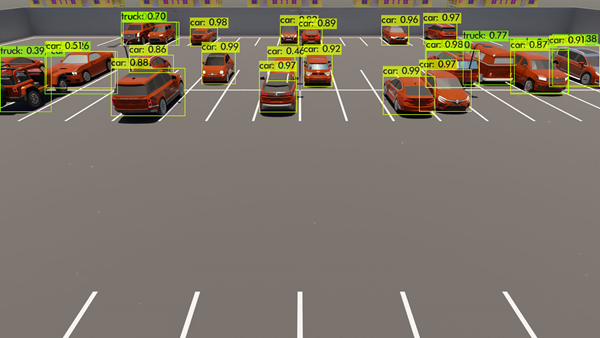} 
  }
  \hspace{-0.25cm}
  \subfloat[\footnotesize Night]{
    \includegraphics[height=3.75cm]{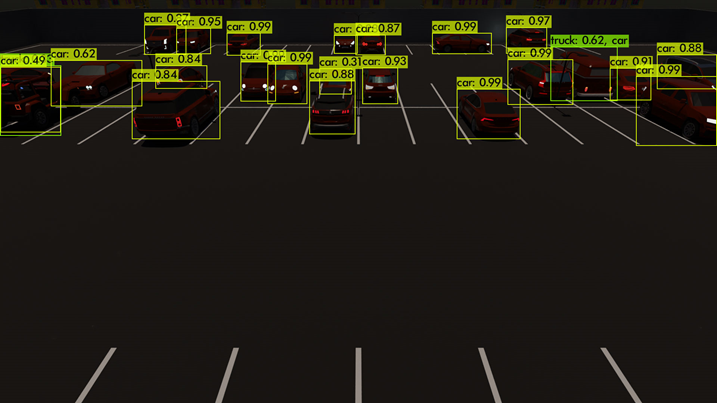}
  }
  \\
  \subfloat[\footnotesize Rain]{
    \includegraphics[height=3.75cm]{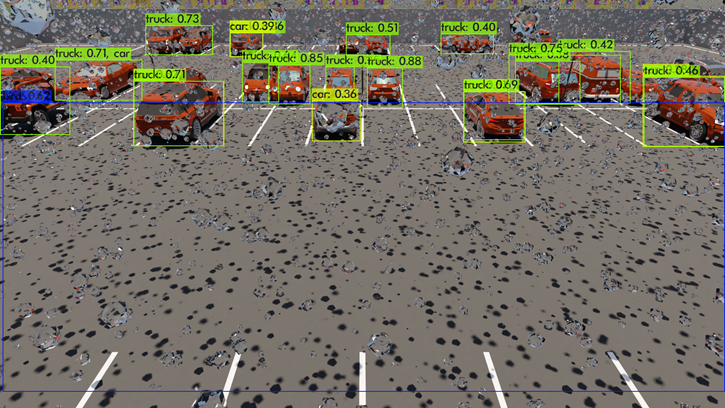} 
  }
  \hspace{-0.25cm}
  \subfloat[\footnotesize Snow]{
    \includegraphics[height=3.75cm]{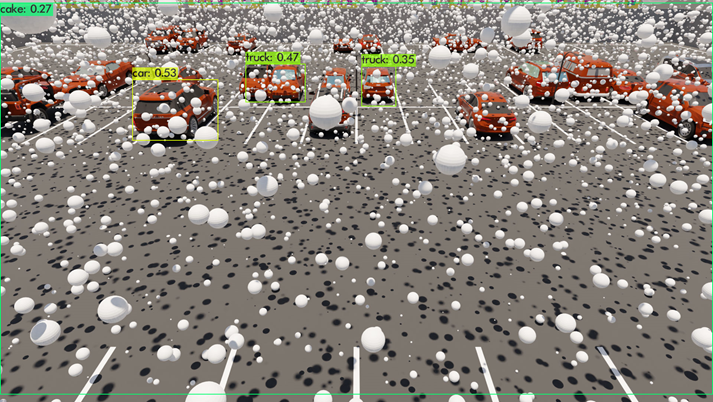}
    \label{fig:maxray_mixed_weather_snow}
  } \\
  \caption{Example of different lighting and weather conditions supported in~\dataset.}
  \label{fig:maxray_mixed_weather}
\end{minipage}
\end{figure}

Further to Sec.~\ref{sec:dataset}, Fig.~\ref{fig:maxray_examples} shows snapshot examples of various data entries from~\dataset.
The examples belong to the parking lot scenario supported in phase 1 of dataset release.
There are currently up to 20 random and identically distributed cars.
The statistics of the dimensions of these cars are depicted in Fig.~\ref{fig:parkingLot_car_stats}.

Fig.~\ref{fig:maxray_mixed_weather} shows examples of different lighting and weather conditions supported in~\dataset. Notice how the reliability of vision detection (Yolo v5 here) drops under unfavourable conditions, particularly snow as depicted in Fig.~\ref{fig:maxray_mixed_weather_snow}.

\newpage

\section{In-depth NNI explanation} \label{sec:nni_appendix}

\begin{table}[b]
\begin{minipage}{\textwidth}
\centering
\scriptsize
\caption{NNI optimisation architecture \& parameters}
\label{tab:NNIparameter}
\begin{tabular}{lllll}
\toprule
\multicolumn{1}{l}{Parameter} & \multicolumn{1}{l}{Explanation}               & \multicolumn{1}{l}{Selection} & \multicolumn{1}{l}{Values}                        & \multicolumn{1}{l}{Best net chosen} \\ 
\cmidrule(lr){1-5}
lr                              & Learning rate                                  & Choice                         & 0.0001, 0.001, 0.01                                & 0.001                                \\
momentum                        & Momentum for optimizer                         & Uniform                        & 0.8, ..., 1                                          & 0.948985588                          \\
act\_func                       & Activation function of conv layer & Choice                         & "ReLU", "LeakyReLU", "Sigmoid", "Tanh", "Softplus" & ReLU                                 \\
optimizer                       & Optimizer type                                 & Choice                         & "SGD", "Adam"                                      & Adam                                 \\
loss\_func                      & Loss function for training only                & Choice                         & "MSE", "L1"                                        & MSE                                  \\
c1\_size                        & Convolutional kernels of c1 layer              & Choice                         & 4, 8, 16, 32, 64                                       & 8                                    \\
c2\_size                        & Convolutional kernels of c2 layer              & Choice                         & 4, 8, 16, 32, 64                                       & 16                                   \\
c3\_size                        & Convolutional kernels of c3 layer              & Choice                         & 4, 8, 16, 32, 64                                       & 8                                    \\
c4\_size                        & Convolutional kernels of c4 layer              & Choice                         & 4, 8, 16, 32, 64                                       & 32                                   \\
k1\_size                        & Kernel size of c1 layer                        & Choice                         & 2, 3, 4                                            & 4                                    \\
k2\_size                        & Kernel size of c2 layer                        & Choice                         & 2, 3, 4                                            & 3                                    \\
k3\_size                        & Kernel size of c3 layer                        & Choice                         & 2, 3, 4                                            & 2                                    \\
k4\_size                        & Kernel size of c4 layer                       & Choice                         & 2, 3, 4                                            & 4                                    \\
s1\_size                        & Stride of c1 layer                             & Choice                         & 1, 2                                                & 2                                    \\
s2\_size                        & Stride of c2 layer                             & Choice                         & 1, 2                                                & 2                                    \\
s3\_size                        & Stride of c3 layer                             & Choice                         & 1, 2                                                & 2                                    \\
s4\_size                        & Stride of c4 layer                             & Choice                         & 1, 2                                                & 1                                    \\
lin1\_size                      & Linear layer 1                                 & Choice                         & 128, 256, 512                                        & 128                                  \\
lin2\_size                      & Linear layer 2                                 & Choice                         & 16, 32, 64, 128, 256                               & 16                                   \\
lin3\_size                      & Linear layer 3                                 & Choice                         & 16, 32, 64, 128, 256                               & 64                                   \\
lin4\_size                      & Linear layer 4                                 & Choice                         & 64, 182, 256                                       & 64       \\
\bottomrule
\end{tabular}
\end{minipage}
\end{table}

Neural Network Intelligence (NNI) is an automatic machine learning (AutoML) tool that enables the systematic exploration of the optimisation space. 
We list the parameters and neural architectures we considered during AutoML optimisation in the Tab.~\ref{tab:NNIparameter}.
The optimal search choice is shown under the right-most column.

\begin{figure}[h]
  \centering
  \includegraphics[width=0.4\textwidth]{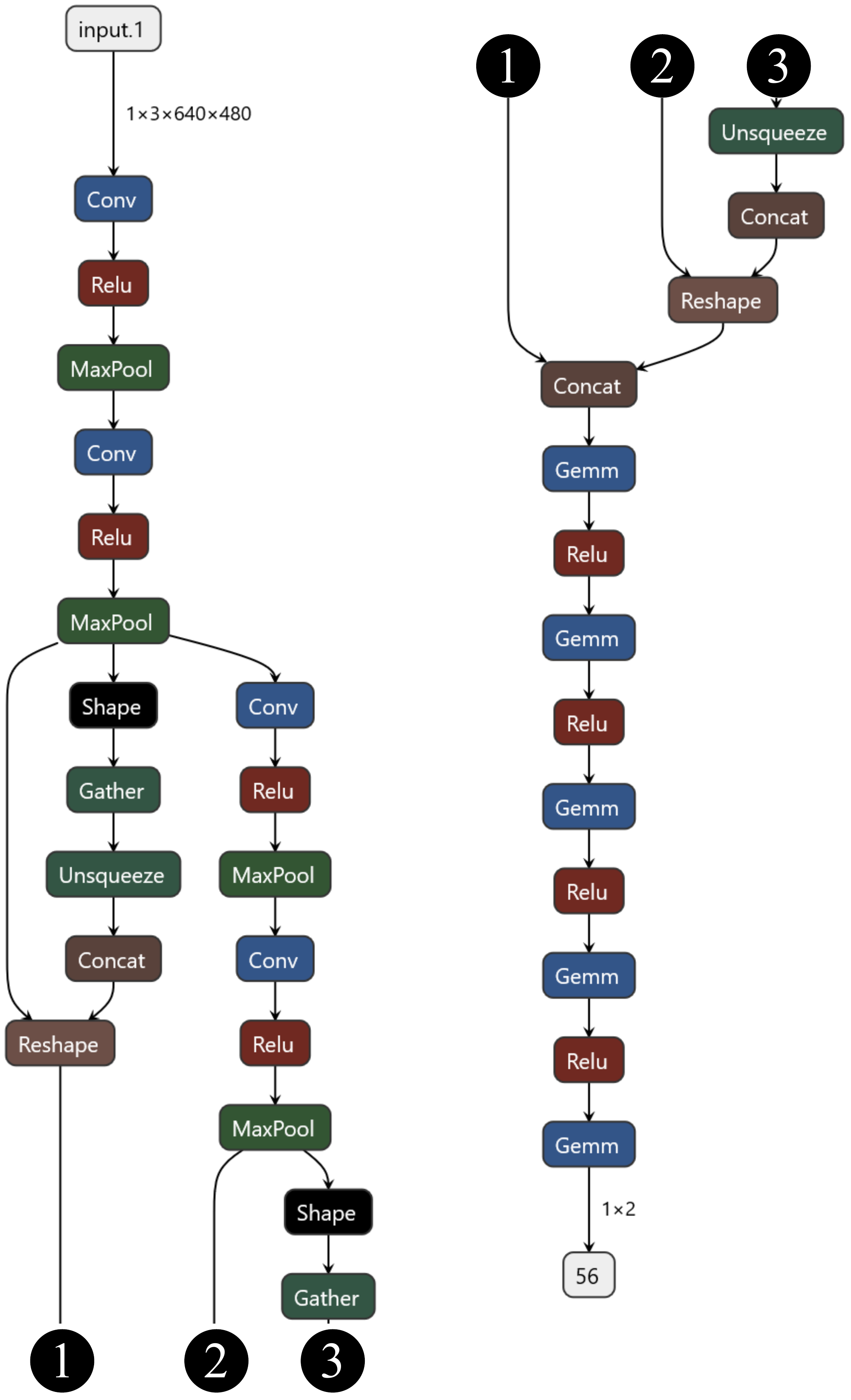}
  \vspace{-0.20cm}
  \caption{Final neural architecture of the supervised localiser network. \protect\circled{1}, \protect\circled{2}, and \protect\circled{3} denote vertical network break and continuation.}
  \label{fig:NNsupervisedStructure}
\end{figure}

Fig.~\ref{fig:NNsupervisedStructure} depicts the final architecture of the supervised network.

\newpage

\vspace{-0.25cm}

\newpage

\section{Dataset comparison} \label{sec:cruw}

Tab.~\ref{tab:rv_datasets_verbose} is a verbose version of Tab.~\ref{tab:rv_datasets_summary} presented in Sec.~\ref{sec:related_work}.
Further, Tab.~\ref{tab:maxray_vs_cruw} summarises the properties of CRUW and how it compares to our~\dataset~dataset.

\begin{table}[b]
\begin{minipage}{\textwidth}
    \vspace{-0.35cm}
    \centering
    \captionsetup{type=table}
    \caption{Radio-visual datasets.}
    \label{tab:rv_datasets_verbose}
    \medskip
  \vspace{-0.50cm}
  \scriptsize
  \setlength{\tabcolsep}{2pt}
  \begin{tabular}{lccccccccccc}
    \toprule
      & \multicolumn{2}{c}{Application} & \multicolumn{3}{c}{Resolution} & \multicolumn{2}{c}{\# of data points}         &             &             &       &                   \\ %
    \cmidrule(lr){2-3} \cmidrule(lr){4-6} \cmidrule(lr){7-8}
    \multicolumn{1}{c}{Dataset}        & Automotive  & 6G            & Range   & Azimuth & Elevation & Total      & Labelled & Frame rate  & Groundtruth & Radar & Reconfigurability \\
    \midrule
    CRUW~\cite{wang2021rodnet}         & \cmark      & \xmark        & 23cm    & 15°     & ---       & 396k$^{1}$ & 260k$^{1}$ & 30          & Camera      & FMCW  & \xmark            \\
    Carrada~\cite{ouaknine2021carrada} & \cmark      & \xmark        & 20cm    & 15°     & ---       &  12.7k     &  7.2k    & 10          & Camera      & FMCW  & \xmark            \\
    AIODrive~\cite{Weng2020_AIODrive}  & \cmark      & \xmark        & N/A     & N/A     & N/A       & 100k       & 100k     & 10          & Synthetic   & N/A   & \xmark            \\
    RADIATE~\cite{sheeny2021radiate}   & \cmark      & \xmark        & 17.5cm  & 1.8°    & 1.8°      & 200k       &  44k     & N/A         & Camera      & FMCW  & \xmark            \\
    Oxford Radar RobotCar~\cite{barnes2020oxford} & \cmark  & \xmark & 4.38cm  & 0.9°    & ---       & 240k       & ---      & 4           & N/A         & FMCW  & \xmark            \\
    RADDet~\cite{zhang2021raddet}      & \xmark      & \cmark        & 19.5cm  & 15°     & 30°       & 10.2k      & 10.2k    & 10          & Camera      & FMCW  & \xmark            \\
    DeepSense~\cite{DeepSense}         & \xmark      & \cmark        & 60cm    & 15°     & 30°       & WIP$^{2}$  & WIP$^{2}$& 10          & Camera+Lidar& FMCW  & \xmark            \\[-0.05cm]
    \midrule \\[-0.4cm]
    \dataset$^{\ast}$                  & \xmark      & \cmark        & 18.75cm & 6.75°   & ---       & 30k        & 30k      & 30          & Synthetic   & OFDM  & \cmark            \\
    \bottomrule             
  \end{tabular}
  \\
  \scriptsize{
  \makebox[14.5cm][l]{$^{1}$only a fraction available publicly.}
  \makebox[14.5cm][l]{$^{2}$work-in-progress: dataset scenarios are being released.}
  \makebox[14.5cm][l]{$^{\ast}$\dataset~is the only 6G synthetic dataset, and is the only reconfigurable dataset.}
  }
\end{minipage}
\end{table}

\begin{table}[b]
\begin{minipage}{\textwidth}
\vspace{0.75cm}
  \caption{Comparison between~\dataset~and CRUW. CRUW$^{\ast}$ requires preprocessing for integration into our radio-visual SSL algorithm.}
  \label{tab:maxray_vs_cruw}
  \scriptsize
  \vspace{-0.25cm}
  \resizebox{\textwidth}{!}{%
  \begin{tabular}{lEEE} 
    \toprule
    Entry     & MaxRay                                       & CRUW                         & Preprocessing  \\
    \cmidrule(lr){1-4}
    Camera    & 30 FPS @ 640$\times$480 pixels               & 30 FPS @ 1440$\times$1080 pixels    & Linear downscaling to 640$\times$480 \\
    Radio     & \multirow{2}{*}{\parbox{3.5cm}{OFDM Radar @ 800MHz BW \\ \underline{dense} 16$\times$16 antenna array}}                       & \multirow{2}{*}{\parbox{3.5cm}{2$\times$ FMCW Radar @ 1250MHz BW \\ \underline{sparse} 4$\times$2 antenna array}}  & \multirow{2}{*}{\parbox{3.5cm}{Radar range filtered to 5-30m, and periodogram upsampled}} \\
              & & & \\
    Range resolution   & 18.75cm                             & 23cm                       &                  \\
    Angular resolution & 6.75°                               & 15°                        &                  \\
    Radio groundtruth  & Perfect high-fidelity ray tracing   & Camera-radar fusion (RODNet labels)    & None \\
    Vision groundtruth & Perfect target bounding box         & Yolov5 target bounding box & None                                          \\
    Scenario      & Parking lot (see Sec.~\ref{sec:dataset}) & Parking lot (see~\cite{wang2021rodnet})      &  \\
     \# of data points & 30k        & 9k      &  \\
    \bottomrule
  \end{tabular}}
  \scriptsize{$^{\ast}$\url{https://www.cruwdataset.org}}
\end{minipage}
\end{table}

\newpage

\onecolumn
\section{Datasheet} \label{sec:datasheet}

We document in Tab.~\ref{tab:dataset_datasheet} various aspects of our radio-visual dataset according to the specifications stipulated in~\cite{gebru2021datasheets}.

\begin{scriptsize}
\begin{longtable}{QA}
\caption{Dataset datasheet}
\label{tab:dataset_datasheet}

\endfirsthead

\toprule
\multicolumn{2}{c}{Cont. Tab.~\ref{tab:dataset_datasheet}}\\
\endhead

\cmidrule(lr){1-2}
\multicolumn{2}{c}{End Tab.~\ref{tab:dataset_datasheet}}\\
\bottomrule
\endlastfoot

\toprule
\multicolumn{2}{c}{Motivation}   \\
\cmidrule(lr){1-2}
For what purpose was the dataset created? 
& To facilitate radio-visual SSL research for 6G sensing. \\
Who created the dataset and on behalf of which entity? 
& Bell Labs Core Research (BLCR) on behalf of Nokia. \\
Who funded the creation of the dataset? 
& Nokia. \\
\cmidrule(lr){1-2}
\multicolumn{2}{c}{Composition}   \\
\cmidrule(lr){1-2}
What do the instances that comprise the dataset represent? 
& heatmap-image pairs sampled from a parking lot scenario. \\
How many instances are there in total? 
& 30,000 labelled for parking lot. \\ 
Does the dataset contain all possible instances or is it a sample of instances from a larger set?  
& All. \\
What data does each instance consist of?  
& Radio heatmaps are range-azimuth description of the environment and RGB images are their visual pairs. \\
Is there a label or target associated with each instance?  
& Object groundtruth coordinates for radio and bounding boxes for vision. \\
Is any information missing from individual instances? 
& No. \\
Are relationships between individual instances made explicit?  
& Correspondence between each radio-visual pair. \\
Are there recommended data splits? 
& 80:20 train-validation split for downstream regression. \\
Are there any errors, sources of noise, or redundancies in the dataset? 
& Not at the data instance level; It is a synthetic dataset. At the radio signal level, high-fidelity propagation modelling captures non-trivial sources of noise such as clutter and fading. \\
Is the dataset self-contained, or does it link to or otherwise rely on external resources? 
& Self-contained. \\
Does the dataset contain data that might be considered confidential? 
& No. \\
Does the dataset contain data that, if viewed directly, might be offensive, insulting, threatening, or might otherwise cause anxiety?  
& No. \\
Does the dataset identify any subpopulations?  
& No. \\
Is it possible to identify individuals, either directly or indirectly from the dataset?  
& No. \\
Does the dataset contain data that might be considered sensitive in any way?  
& No. \\
\cmidrule(lr){1-2}
\multicolumn{2}{c}{Collection Process}   \\
\cmidrule(lr){1-2}
How was the data associated with each instance acquired?  
& Synthesised using CAD tools. \\
What mechanisms or procedures were used to collect the data?  
& Ray-tracing for radio and rendering for vision. \\
If the dataset is a sample from a larger set, what was the sampling strategy? 
& N/A. \\
Who was involved in the data collection process and how were they compensated?  
& Nokia employees under full-time employment. \\
Over what timeframe was the data collected?  
& Data generation took several months of in-house development effort. \\
Were any ethical review processes conducted?  
& N/A. \\
Did you collect the data from the individuals in question directly, or obtain it via third parties or other sources?  
& N/A. \\
Were the individuals in question notified about the data collection?  
& N/A. \\
Did the individuals in question consent to the collection and use of their data?  
& N/A. \\
If consent was obtained, were the consenting individuals provided with a mechanism to revoke their consent in the future or for certain uses?  
& N/A. \\
Has an analysis of the potential impact of the dataset and its use on data subjects been conducted?  
& N/A. \\
\cmidrule(lr){1-2}
\multicolumn{2}{c}{Preprocessing/cleaning/labeling}   \\
\cmidrule(lr){1-2}
Was any preprocessing/cleaning/labeling of the data done?  
& No. \\
Was the ``raw'' data saved in addition to the preprocessed/cleaned/labeled data?  
& N/A. \\
Is the software that was used to preprocess/clean/label the data available?  
& N/A. \\
\cmidrule(lr){1-2}
\multicolumn{2}{c}{Uses}   \\
\cmidrule(lr){1-2}
Has the dataset been used for any tasks already? 
& Mainly radio-visual SSL research disclosed in this paper. \\
Is there a repository that links to any or all papers or systems that use the dataset?  
& N/A. \\
What (other) tasks could the dataset be used for?  
& This is a 1st radio-visual SSL work, and future research would build on our ideas and/or investigate alternative approaches, e.g., for more discriminative radio signals obtained from finer angular resolutions. \\
Is there anything about the composition of the dataset or the way it was collected and preprocessed/cleaned/labeled that might impact future uses?  
& No. \\
Are there tasks for which the dataset should not be used?  
& N/A. \\
\cmidrule(lr){1-2}
\multicolumn{2}{c}{Distribution}   \\
\cmidrule(lr){1-2}
Will the dataset be distributed to third parties outside of the entity on behalf of which the dataset was created?  
& Yes. \\
How will the dataset will be distributed?  
& Hosted on a public website. \\
When will the dataset be distributed?  
& 2023. \\
Will the dataset be distributed under a copyright or other intellectual property (IP) license, and/or under applicable terms of use (ToU)?  
& Yes. \\
Have any third parties imposed IP-based or other restrictions on the data associated with the instances?  
& No. \\
Do any export controls or other regulatory restrictions apply to the dataset or to individual instances?  
& No. \\
\cmidrule(lr){1-2}
\multicolumn{2}{c}{Maintenance}   \\
\cmidrule(lr){1-2}
Who will be supporting/hosting/maintaining the dataset?  
& Nokia Bell Labs. \\
How can the owner/curator/manager of the dataset be contacted?  
& Email. \\
Is there an erratum?  
& No. \\
Will the dataset be updated?  
& Yes. \\
If the dataset relates to people, are there applicable limits on the retention of the data associated with the instances?  
& N/A. \\
Will older versions of the dataset continue to be supported/hosted/maintained?  
& Yes. \\
If others want to extend/augment/build on/contribute to the dataset, is there a mechanism for them to do so?  
& We will provide reference Blender files which can be modified to model different environments.
Our radio raytracing is however proprietary and cannot be released.
To work around this, users could licence equivalent commercial radio raytracers in order to generate paired radio heatmaps from Blender's 3D models. \\
\end{longtable}
\end{scriptsize}
\twocolumn

\end{appendices}

\end{document}